\documentclass{article}

\usepackage[final]{neurips_2025}

\usepackage[utf8]{inputenc} 
\usepackage[T1]{fontenc}    
\usepackage{url}   

\usepackage[pagebackref=true,breaklinks=true,letterpaper=true,colorlinks,bookmarks=false,citecolor=blue]{hyperref}

\usepackage{booktabs}       
\usepackage{amsfonts}       
\usepackage{nicefrac}       
\usepackage{microtype}      
\usepackage{xcolor}         
\usepackage{colortbl}
\usepackage{subcaption}
\usepackage{enumitem}
\usepackage{graphicx}
\usepackage{multirow}
\usepackage{amsmath}
\usepackage{makecell}
\usepackage{amssymb}
\usepackage{float}
\usepackage{pifont}       
\usepackage{bbding}       
\usepackage{fontawesome}  
\usepackage{wrapfig}
\usepackage{tikz}
\usepackage{tcolorbox}
\usepackage{soul}
\usepackage{tabularx}
\usepackage[ruled,vlined]{algorithm2e}
\usepackage{algpseudocode}
\usepackage{amsthm}

\newtheorem{hypothesis}{Hypothesis}
\newtheorem{theorem}{Theorem}[section] 
\newtheorem{proposition}[theorem]{Proposition} 

\title{AdaLRS: Loss-Guided Adaptive Learning Rate Search for Efficient Foundation Model Pretraining}

\author{%
    Hongyuan Dong, Dingkang Yang, Xiao Liang, Chao Feng\textsuperscript{\dag}, Jiao Ran \\
    ByteDance Inc. \\
    \texttt{d\_ousia@icloud.com, yangdingkang@bytedance.com,  liangxiao.ilx@bytedance.com} \\
    \texttt{chaofeng.zz@bytedance.com, ranjiao@bytedance.com} \\
}

\begin{document}

\maketitle
\newcommand\blfootnote[1]{%
    \begingroup
    \renewcommand\thefootnote{}\footnote{#1}%
    \addtocounter{footnote}{-1}%
    \endgroup
}
\blfootnote{\dag Email corresponding}

\begin{abstract}
Learning rate is widely regarded as crucial for effective foundation model pretraining.
Recent research explores and demonstrates the transferability of learning rate configurations across varying model and dataset sizes, etc. 
Nevertheless, these approaches are constrained to specific training scenarios and typically necessitate extensive hyperparameter tuning on proxy models.
In this work, we propose \textbf{AdaLRS}, a plug-in-and-play adaptive learning rate search algorithm that conducts online optimal learning rate search via optimizing loss descent velocities.
We provide theoretical and experimental analyzes to show that foundation model pretraining loss and its descent velocity are both convex and share the same optimal learning rate. 
Relying solely on training loss dynamics, AdaLRS involves few extra computations to guide the search process, and its convergence is guaranteed via theoretical analysis. 
Experiments on both LLM and VLM pretraining show that AdaLRS adjusts suboptimal learning rates to the neighborhood of optimum with marked efficiency and effectiveness, with model performance improved accordingly. 
We also show the robust generalizability of AdaLRS across varying training scenarios, such as different model sizes, training paradigms, base learning rate scheduler choices, and hyperparameter settings.
\end{abstract}

\vspace{-5pt}
\section{Introduction}
\vspace{-5pt}

Learning rate (LR) is regarded as a critical hyperparameter for foundation model pretraining, \textit{e.g.} Large Language Model (LLM), Vision Language Model (VLM), etc. 
With the development of foundation model pretraining techniques~\cite{achiam2023gpt,liu2024deepseek,yang2024qwen2,yao2024minicpm,chen2024far,chen2024expanding,dong2025scalable,lei2025scalability}, a series of studies seek to find the optimal learning rate for effective pretraining~\cite{shen2024power,yang2022tensor,li2025predictable,mckinzie2024mm1}.
Although reducing the costs of hyperparameter searches to some extent, these methods are bound to specific pretraining scenarios or necessitate resource-consuming hyperparameter searches on proxy models to take effect. 

One line of work predicts the optimal learning rate settings directly via summarizing model performance dynamics \textit{w.r.t.} hyperparameter settings.
Under certain model structure designs, the optimal learning rate can be expressed as a function of a series of training hyperparameters~\cite{kaplan2020scaling,bi2024deepseek,li2025predictable,mckinzie2024mm1}, such as model size, batch size, training budget, etc.
Nevertheless, the resulting correlation laws lack desired generalizability to new training scenarios, where large-scale learning rate search experiments are required to establish the model performance dynamic patterns.

Another line of research explores transferring hyperparameter search results obtained from proxy models to larger ones. 
Tensor Program series work~\cite{yang2021tensor,yang2022tensor} proves the zero-shot transferability of hyperparameter settings across model sizes for transformer models.
As long as the model design meets certain criteria, learning rate search on certain proxy models can be transferred to larger models directly, reducing the hyperparameter search cost significantly. 
However, the learning search process on proxy models is still time- and resource-consuming. 
Although auto hyperparameter learning techniques can be applied to accelerate the search process~\cite{bergstra2012random,snoek2012practical,li2018hyperband}, a large number of independent runs are still required to approximate the optimum, resulting in additional training resource overhead. 

In this work, we propose AdaLRS, a plug-in-and-play learning rate search algorithm that optimizes the loss descent velocity to find the optimal learning rate in a single run. 
We formulate the optimization of training loss and loss descent velocity w.r.t LR in foundation model pretraining tasks as convex problems, and show that they share the same optimum by experiment results. 
Based on these observations, AdaLRS adjusts the base learning rate sequentially according to training loss dynamics to approximate the optimum. 
We provide theoretical proof for the convergence and geometric error decay of AdaLRS, and show that AdaLRS effectively tackles various foundation model pretraining tasks.
It adjusts inappropriate LRs effectively in single runs, and improves model performance significantly for different model sizes, training paradigms, and base scheduler choices.

To summarize, our contributions can be listed as follows:
\begin{itemize}[leftmargin=*]
\item We provide theoretical and experimental analysis to demonstrate that the foundation model pretraining loss and its slop \textit{w.r.t.} LR are convex and share the same optimum. 

\item We propose AdaLRS, an learning rate search algorithm which performs online LR adjustment to optimize loss descent velocity, approximating the optimal learning rate in a single run. 
We provide theoretical analysis for its convergence and the geometric error decay in the search process. 

\item  We conduct experiments in both LLM and VLM pretraining tasks with varying base learning rates.
AdaLRS not only demonstrates marked effectiveness in finding the optimal LR and improving model performance, but also exhibits desired generalizability in different pretraining tasks. 

\end{itemize}



\vspace{-5pt}
\section{AdaLRS}
\vspace{-5pt}
In this section, we provide formulations for the proposed AdaLRS algorithm and present proof of its convergence and geometric error decay in the optimal LR search process. 

\subsection{Formulations of AdaLRS Algorithm}
Previous work has discussed the convexity of training loss optimization \textit{w.r.t.} learning rate settings~\cite{schaipp2025surprising}. 
In this work, we further hypothesize that the optimization of the training loss and loss curve slope in foundation model pretraining tasks are both near-convex and share the same optimal learning rate, which is supported by the experiment results shown in Section~\ref{sec: 3.2}. 
In this way, we can approximate the optimal learning rate by optimizing the velocity of loss descent within a single run.

\textbf{Formulation. }
Denoting the learning rate at training step $t$ as $\eta_t$, and the LR upscaling and downscaling factors as $\alpha'$ and $\beta'$, we formulate the workflow of AdaLRS as follows: 

\begin{equation}
\eta_{t+k} = 
\begin{cases} 
\alpha' \eta_t & \text{if } v(\alpha'\eta_t) > v(\eta_t) + 2e \quad (\text{loss slope \textit{increases} $\uparrow$}), \\
\beta' \eta_t & \text{if } v(\alpha'\eta_t) < v(\eta_t) - 2e \quad (\text{loss slope \textit{decreases} $\downarrow$}), \\
\eta_t & \text{otherwise.}
\end{cases}
\label{eq: main}
\end{equation}

$v(\cdot)$ indicates the estimated loss curve slope obtained from a $k$-step window, while $e$ stands for the estimation error between $v$ and the true loss descent velocity $V$.
$\alpha'$ and $\beta'$ are rectified LR scaling factors which satisfy $\alpha'=\max(\lambda^t\alpha, 1), \beta'=\frac{1}{\max(\lambda^t\beta, 1)}$, where $\alpha, \beta > 1$ are two multiplicatively independent real numbers and $\lambda \in (0, 1)$ is a decay factor.
We validate the multiplicatively independent design of LR scaling factors (for all integers $m,n$, $\alpha^m=\beta^n \implies m=n=0$) in Appendix~\ref{appendix:val}.

\textbf{Workflow. }
\label{sec:adalrs_workflow}
During model training, AdaLRS monitors the loss curve slope $v_t$ with the least squares method~\cite{bjorck1990least}, and attempts to upscale the learning rate when the loss curve slope decays.
After the upscaling adjustment, we compare the loss curve slope with that before upscaling. 
As shown in Equation~\ref{eq: main}, if the estimated loss slope increases more than $2e$ after the adjustment, the upscaling is regarded as valid and the adjustment will be retained. 
On the other hand, once the validation fails under condition $V(\beta'\eta_t) < V(\eta_t) - 2e$, the upscaling adjustment is reverted, with a downscaling factor applied to the base learning rate. 
As a result, AdaLRS is able to conduct an online search for the optimal learning rate within a single run, performing marked efficiency and effectiveness. 

We also introduce several tricks to improve the stability of the AdaLRS algorithm. 
In the LR upscaling process, we adopt an early stopping operation if the training loss rises over the largest loss value in the loss records, followed by several trial training steps performed after upscaling. 
This is to ensure that the loss slopes before and after the trial adjustment can be compared fairly at similar loss levels. 
Additionally, the learning rate search process is restricted to certain step ratios in the early training stage, ensuring a stable LR decay process for pretraining. 
To tackle extremely high learning rate settings, we also introduce a boundary condition that lowers the LR if the loss increases for two consecutive windows.
We refer to Appendix~\ref{appendix:adalrs_algo} for a detailed formulation of AdaLRS in pseudocode. 


\subsection{Convergence Analysis}
\label{sec: 3.2}

\subsubsection{Theoretical Hypotheses}
We make two hypotheses to prove the convergence of the proposed AdaLRS algorithm. 

\begin{hypothesis}
\label{hypo:1}
The optimization of the training loss \textit{w.r.t.} learning rate in foundation model pretraining is convex, and share the same optimum with loss descent velocity optimization. 
Let $V(\eta)$ and $L(\eta)$ be the loss descent velocity and loss value function \textit{w.r.t.} the learning rate $\eta$, this hypothesis means that there exists an optimal learning rate $\eta^*$, so that: 
\begin{equation}
\begin{cases}
    \frac{\partial V}{\partial \eta} > 0 \ \text{and} \ \frac{\partial L}{\partial \eta} < 0, & \forall \eta < \eta^*, \\
    \frac{\partial V}{\partial \eta} < 0 \ \text{and} \ \frac{\partial L}{\partial \eta} > 0, & \forall \eta > \eta^*.
\end{cases}
\end{equation}
\end{hypothesis}

This is a relatively strong assumption, and therefore we conduct both theoretical and experimental analyses to support it.
Consider a simplified foundation model pretraining task with a constant learning rate and SGD optimizer, we formulate the update rule for model parameter $\psi$ as:
\begin{equation}
    \psi_{t+1} = \psi_t - \eta \nabla L_t, 
\end{equation}
where $L_t$ is the training loss at time step $t$. 
The expected loss descent velocity per step (using Taylor expansion) is:
\begin{equation}
    \mathbb{E}\left[ L_{t+1} - L_t \right] 
    \approx -\eta \| \nabla L_t \|^2 + \frac{C_{Lip}}{2} \eta^2 \| \nabla L_t \|^2, 
\end{equation}
where $C_{Lip}$ is the Lipschitz constant of the gradient.
When the learning rate is small, the first term dominates the expectation, and a smaller $\eta$ leads to a smaller decrease in loss. When the learning rate is large, on the other hand, the second term cannot be neglected. It contributes to suppress the expected loss descent value. 
Differentiating w.r.t. $\eta$ and setting to zero for the extreme:
\begin{equation}
    \frac{\partial}{\partial \eta} \left( -\eta + \frac{C_{Lip}}{2} \eta^2 \right) = 0 \implies \eta^* = \frac{1}{C_{Lip}}.
\end{equation}
This shows that the loss descent velocity w.r.t. $\eta$ is a convex function.
It is intuitive to infer that a higher loss descent velocity leads to a lower training loss, and vice versa, which proves that the optimization of training loss is also a convex problem.

\begin{figure}[thb]
\vspace{-3pt}
    \centering
    \begin{subfigure}{0.31\textwidth}
        \includegraphics[width=\textwidth]{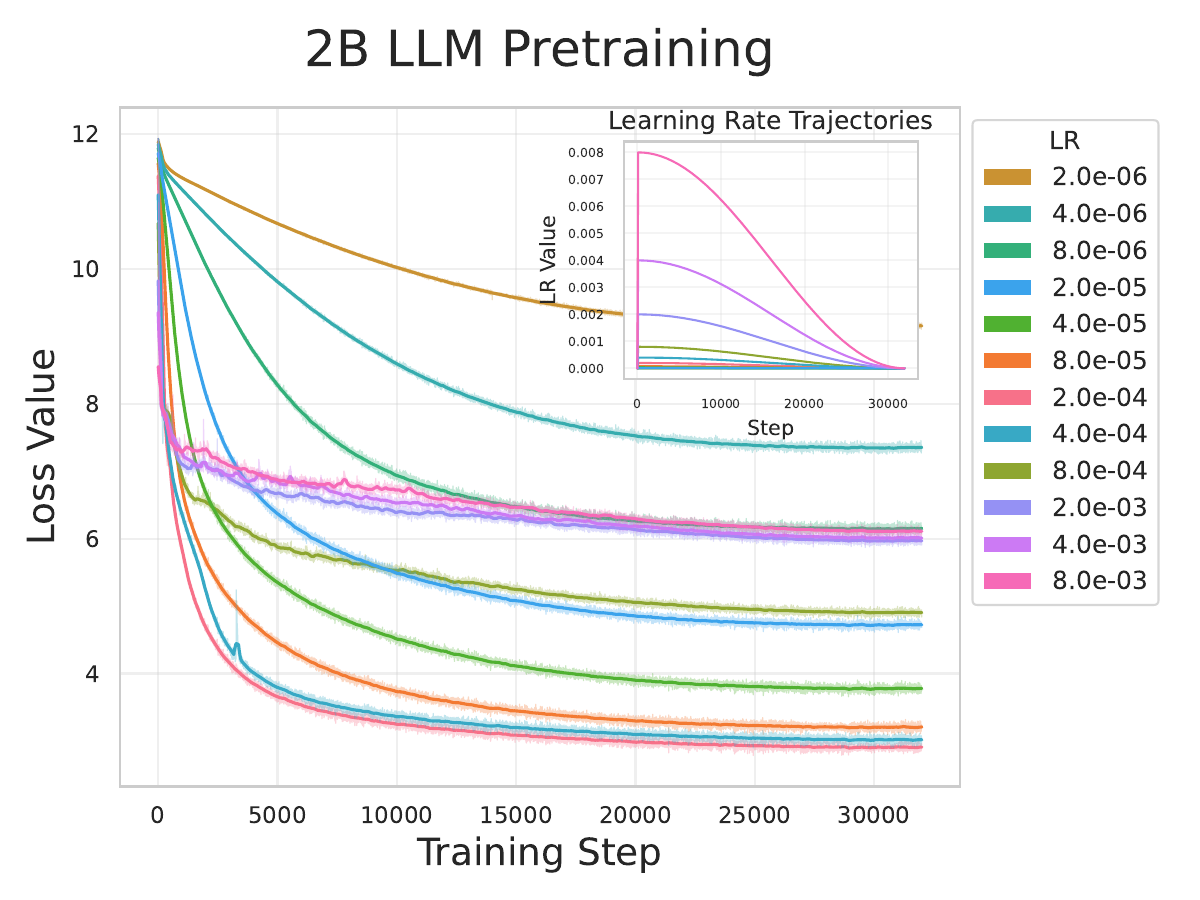}
        \caption{}
    \end{subfigure}
    \begin{subfigure}{0.31\textwidth}
        \includegraphics[width=\textwidth]{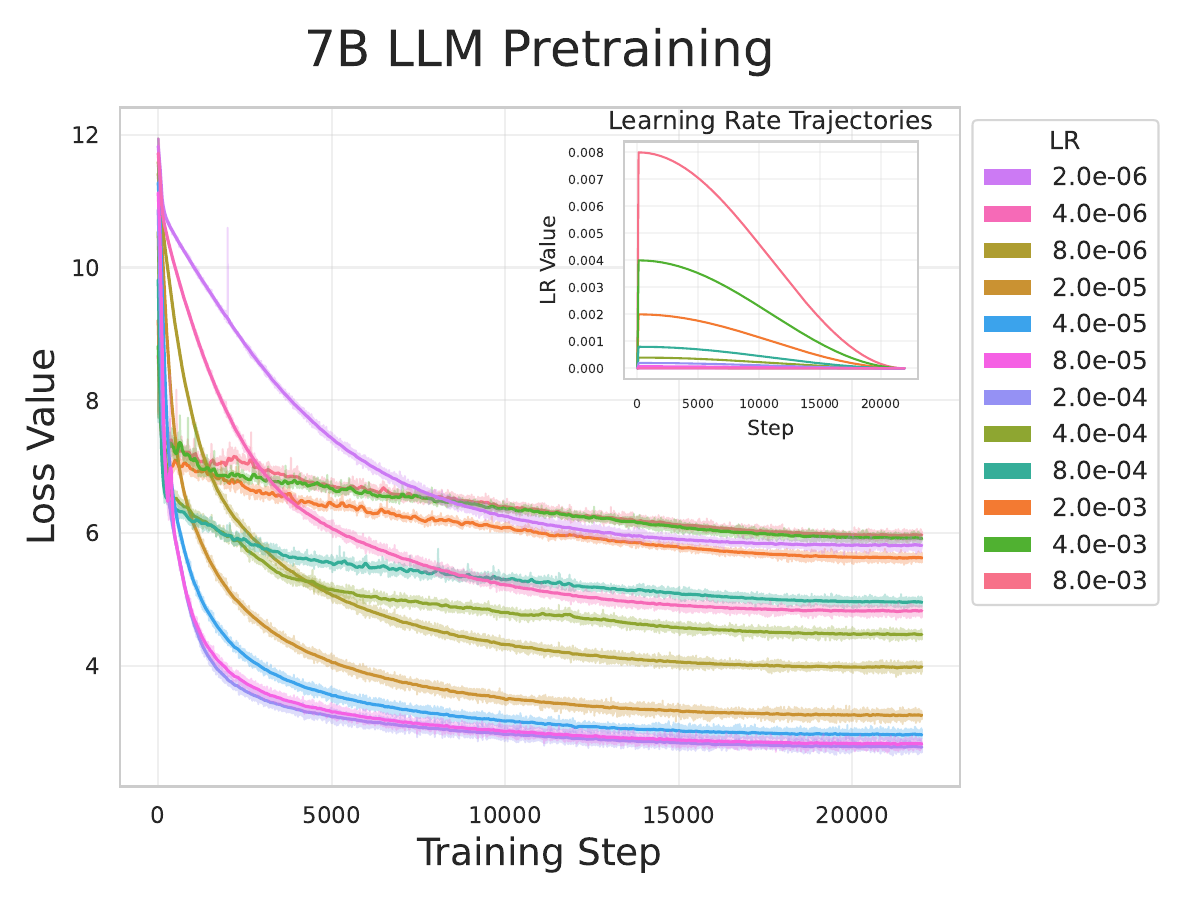}
        \caption{}
    \end{subfigure}
    \begin{subfigure}{0.31\textwidth}
        \includegraphics[width=\textwidth]{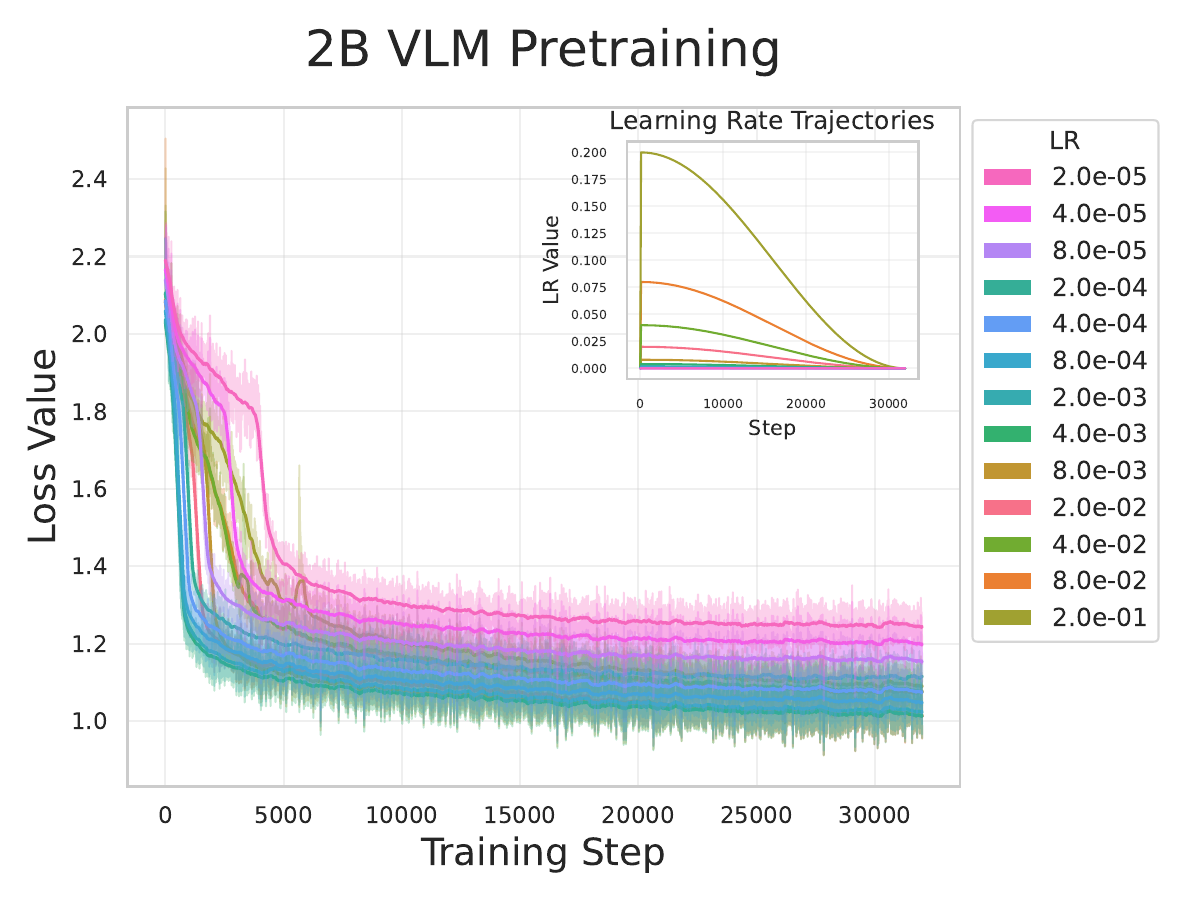}
        \caption{}
    \end{subfigure}

    \begin{subfigure}{0.31\textwidth}
        \includegraphics[width=\textwidth]{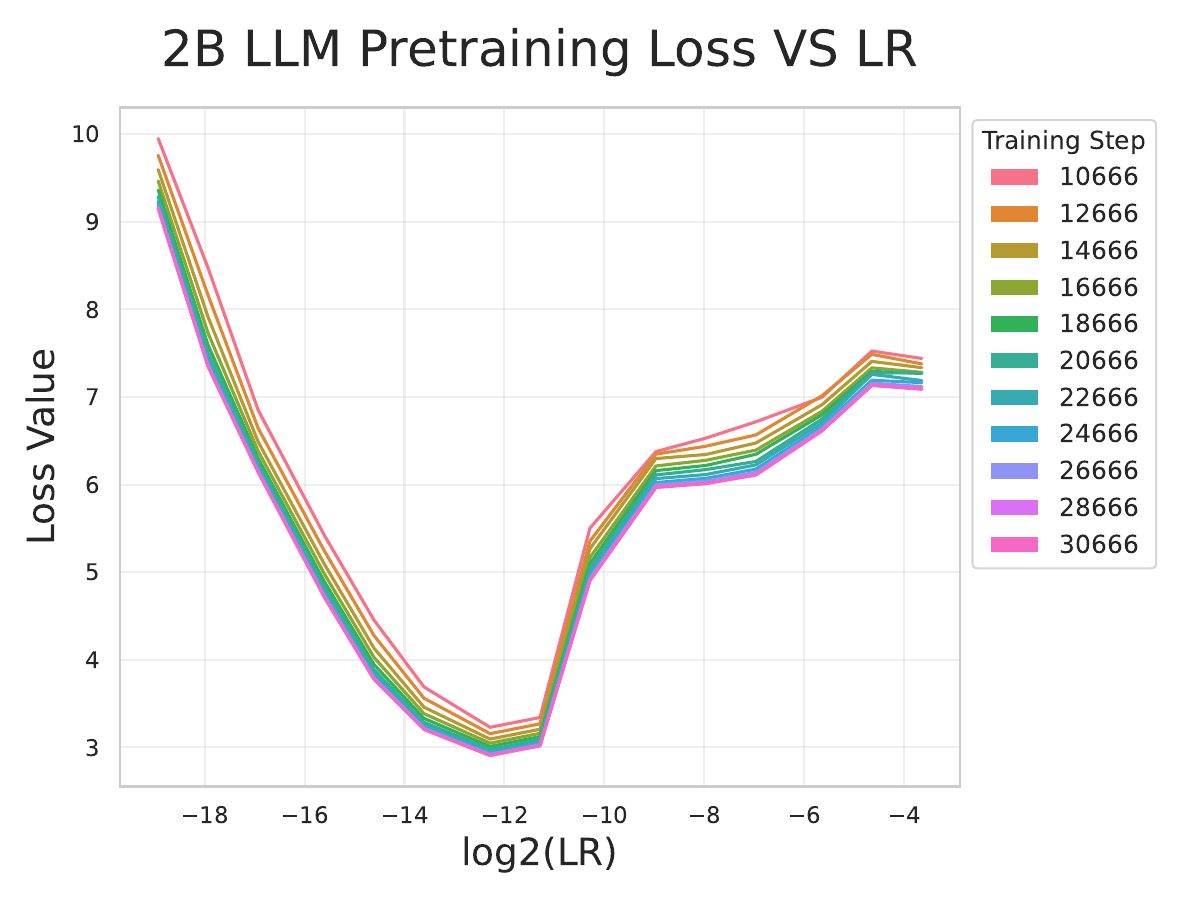}
        \caption{}
    \end{subfigure}
    \begin{subfigure}{0.31\textwidth}
        \includegraphics[width=\textwidth]{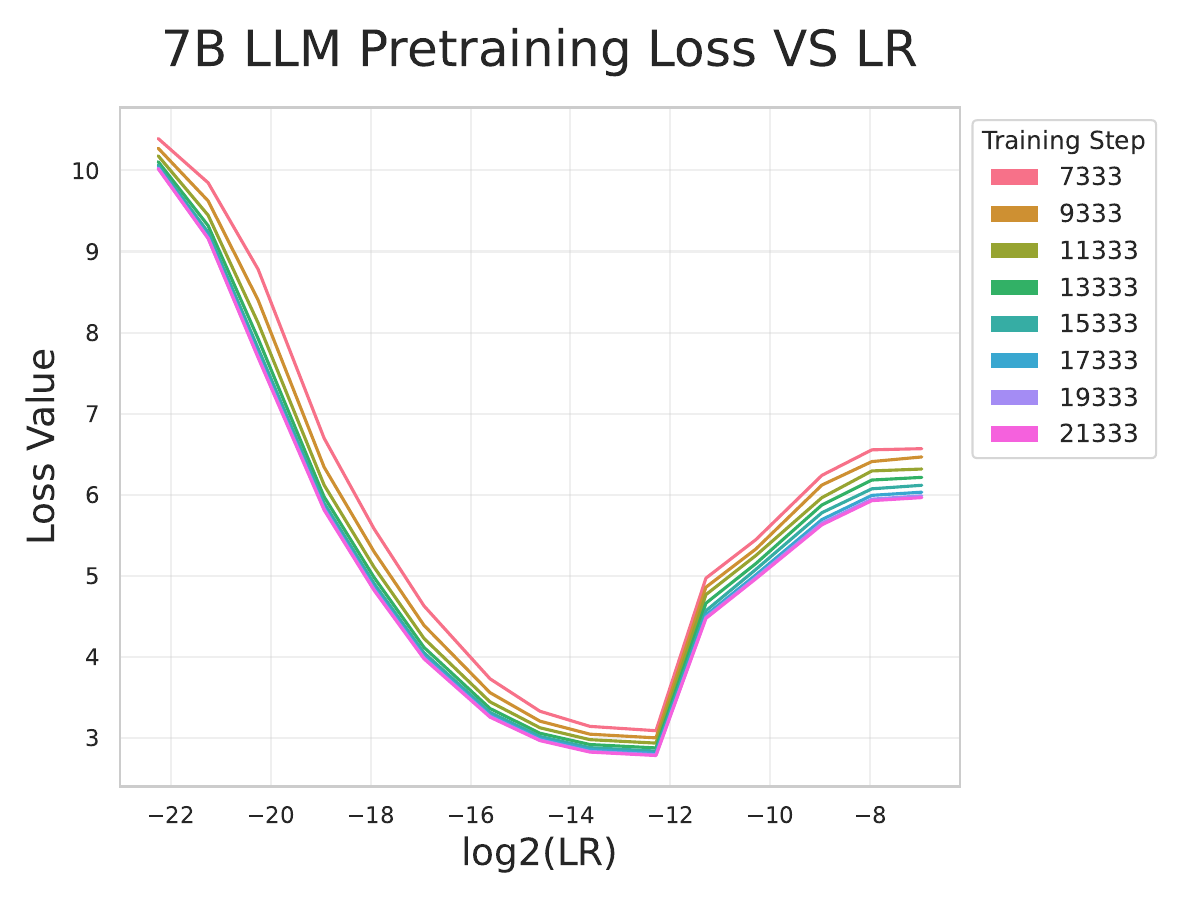}
        \caption{}
    \end{subfigure}
    \begin{subfigure}{0.31\textwidth}
        \includegraphics[width=\textwidth]{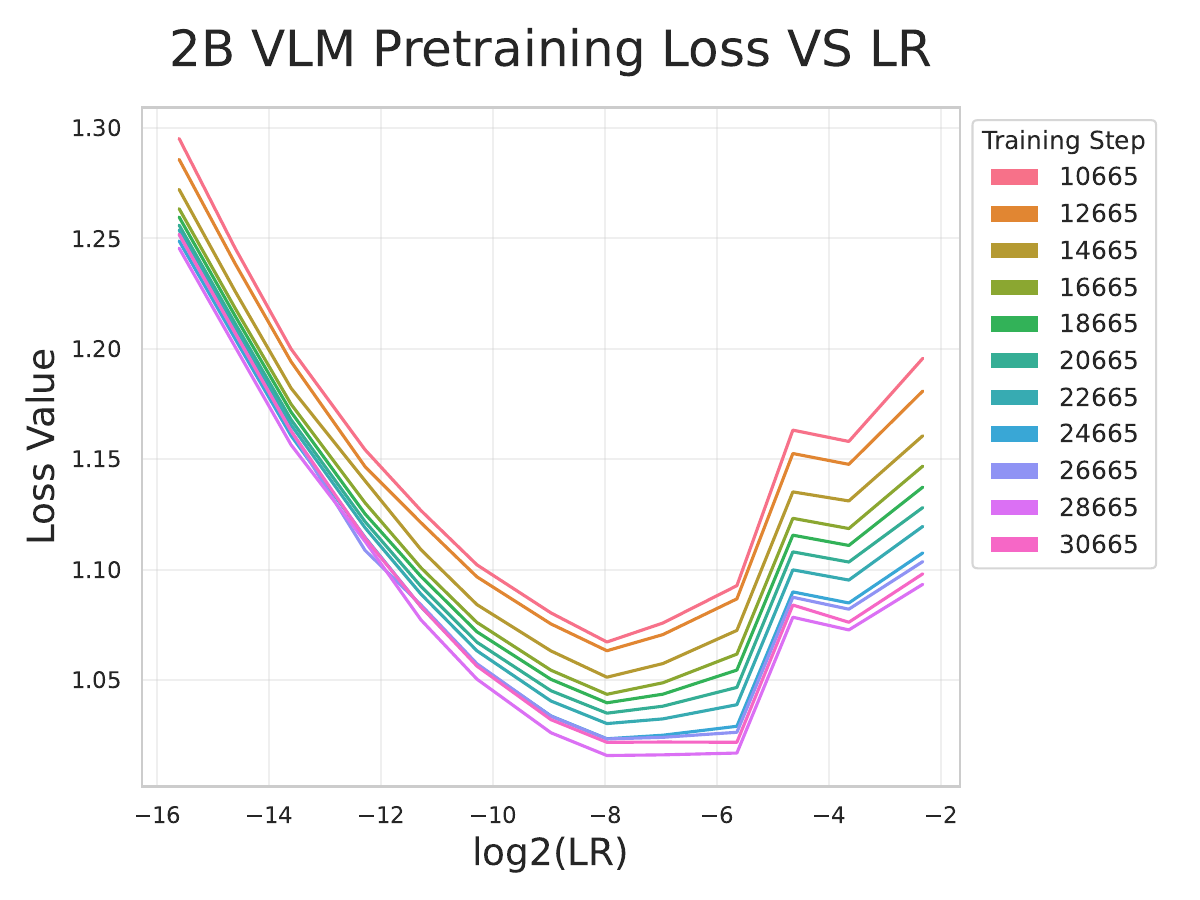}
        \caption{}
    \end{subfigure}

    \begin{subfigure}{0.31\textwidth}
        \includegraphics[width=\textwidth]{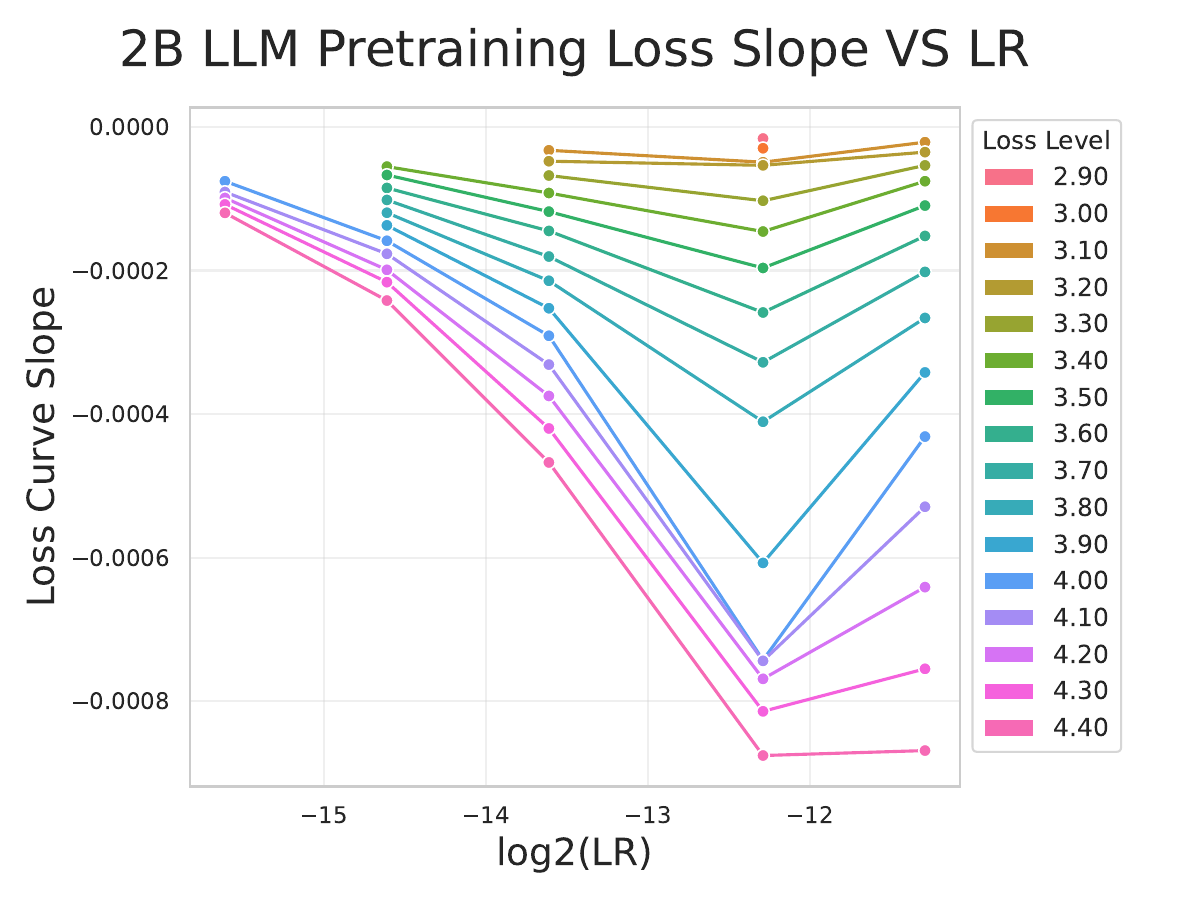}
        \caption{}
    \end{subfigure}
    \begin{subfigure}{0.31\textwidth}
        \includegraphics[width=\textwidth]{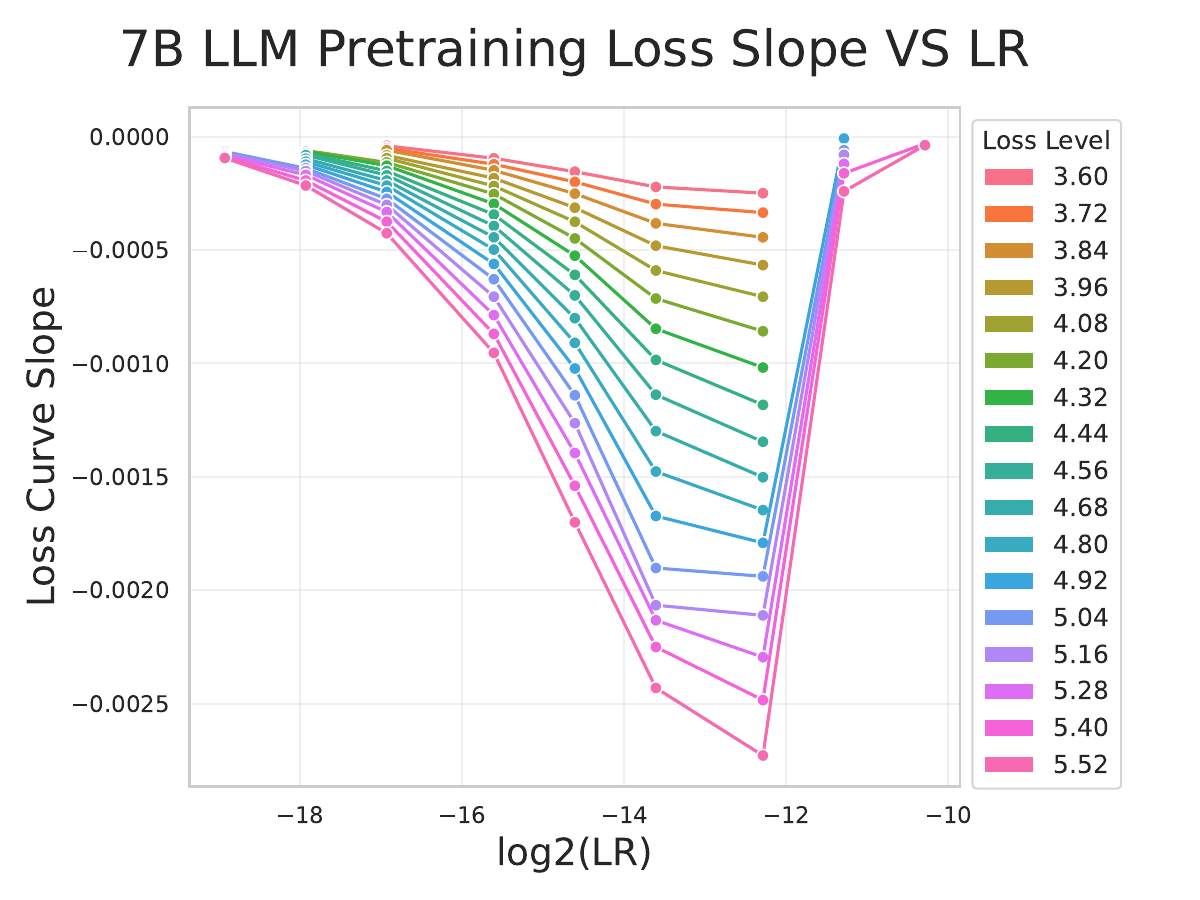}
        \caption{}
    \end{subfigure}
    \begin{subfigure}{0.31\textwidth}
        \includegraphics[width=\textwidth]{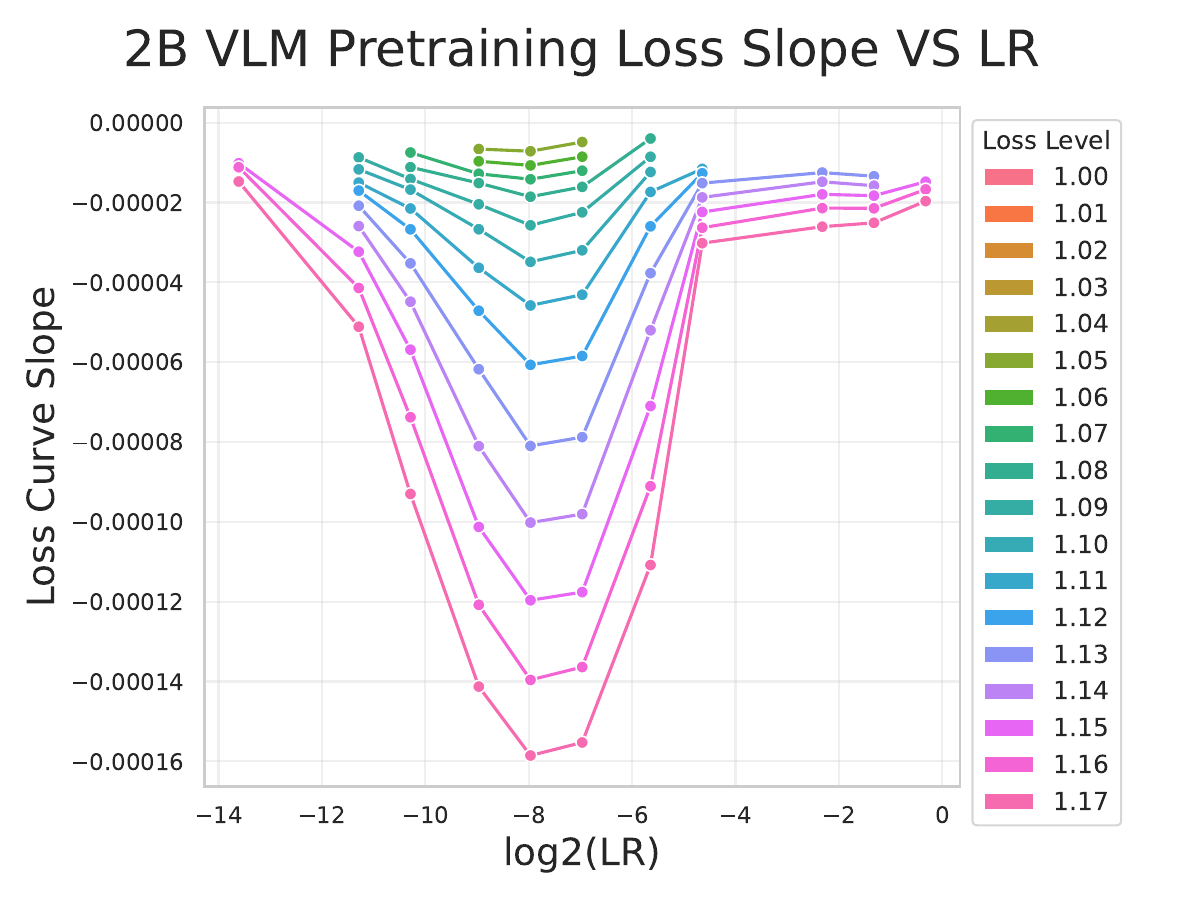}
        \caption{}
    \end{subfigure}
    
    \caption{
        Training loss and loss descent velocity dynamics \textit{w.r.t.} varying LR settings for LLM and VLM pretraining. 
        Figures (a)\&(b)\&(c) show the training losses and LR trajectories with a cosine LRS, while Figures (d)\&(e)\&(f) illustrate how training loss varies across different LR settings through the training process. 
        Figures (g)\&(h)\&(i) are loss slope dynamics at varying loss levels, obtained from experiments with constant learning rates. 
    }
    \label{fig:pilot}
\end{figure}

Apart from theoretical analyses, we also conduct small-scale pretraining experiments in LLM and VLM pretraining to verify this hypothesis. 
For LLM pretraining, we use Qwen2.5-1.5B and Qwen2.5-7B~\cite{qwen2.5} models with randomly initialized parameters, and train them from scratch on 8M samples from the SlimPajama~\cite{cerebras2023slimpajama} dataset. 
A series of exponentially larger learning rates in $[2e^{-6},8e^{-3}]$ are applied to investigate the dynamics of training loss and loss curve slope under varying LR settings.
For VLM pretraining, we conduct experiments on a 2B model with the architecture described in SAIL-VL~\cite{dong2025scalable}. 
Approximately 8M data consisting of caption and OCR samples are used for model training, with learning rates configured in $[2e^{-5},2]$. 

The resulted training loss curves are shown in Figures~\ref{fig:pilot}(a)(b)(c), with training loss and loss descent velocity dynamics \textit{w.r.t.} varying LR settings illustrated in Figures~\ref{fig:pilot}(d)(e)(f) and (g)(h)(i), respectively.
As Figures~\ref{fig:pilot}(d)(e)(f) shows, pretraining losses at different training steps exhibit strong correlations with learning rate settings, forming a series of convex curves with constant optima. 
To investigate how learning rate influences the loss descent velocity, we conduct several sets of experiments with constant LR values. 
We show the correlation between loss curve slopes and learning rate settings in Figures~\ref{fig:pilot}(g)(h)(i), and the corresponding curves are not only convex, but also sharing the same optima as the training loss curves.
Some data points in Figures~\ref{fig:pilot}(g)(h)(i) are not displayed because the corresponding inappropriate learning rate cannot optimize the model to such loss levels. 
These experiment results strongly support the validity of Hypothesis~\ref{hypo:1}.

In practice, $V$ is also influenced by the current loss value and optimization states. 
As described in Section~\ref{sec:adalrs_workflow}, AdaLRS performs trial training steps before comparing loss slopes to eliminate the influence of loss levels, and we assume that the difference in optimization states is limited between consecutive training steps. 
As a result, we present $V$ as a univariate function in our proof. 


\begin{hypothesis}
\label{hypo:2}
Let $v_t$ denote the empirical estimate of $V(\eta_t)$, computed over a sliding window of $k$ training steps. 
Denoting the estimation error bound as $e$, i.e. $\forall t>0,\, |v_t-V(\eta_t)|<e$, there exists a sliding window length $k$ such that $e$ can be sufficiently small. 
\end{hypothesis}
This guarantees that AdaLRS converges to a relatively narrow neighborhood of the optimum $\eta^*$. 

\subsubsection{Convergence}
Equipped with the above hypotheses, we seek to prove the following theorem: 
\begin{theorem}
\label{theorem:1}
The learning rate sequence $\{\eta_t\}$ generated by the algorithm converges almost surely to the $e$-neighborhood of the optimal learning rate \(N_e(\eta^*) \triangleq \{\eta : |\eta - \eta^*| < e\}\), i.e.,
\begin{equation}
    \mathop{\text{lim}}\limits_{t \rightarrow \infty} \mathbb{P}(|\eta_t - \eta^*| < e) = 1.
\end{equation}
\end{theorem}

\begin{proposition}
\label{prop:1}
    If \(\eta_t, \max(\lambda^t\alpha, 1)\eta_t < \eta^*-e\), \(\exists T < \infty\) such that \(\eta_{t+T}\) moves towards \(\eta^*\), and vice versa. 
\end{proposition}

Since we gradually push the the loss slope decay threshold $\theta$ to $1$ (detailed in Appedix~\ref{appendix:adalrs_algo}), $v_{t+k} < v_t * \theta$ will almost surely trigger the LR upscaling adjustment in finite steps. 
We denote the time step as $T$, with the loss descent velocity before or after trial LR adjustment as $v_T$ and $v_{T+k}$, respectively.
The LR upscaling adjustment is kept and only kept when $v_{T+k}-v_T>2e$, which can be reformulated as $V(\eta_{T+k})+e_{T+k}-V(\eta_T)-e_T > 2e$, where $e_{T+k}$ and $e_T$ stand for loss slope estimation error at corresponding steps. 
Substituting gives $V(\eta_{T+k})-V(\eta_T) > 2e-e_{T+k}+e_T > 2e-|e_{T+k}|-|e_T| > 0$, which means the upscaling condition in Equation~\ref{eq: main} ensures an increase in the true loss descent velocity. 
According to Hypothesis~\ref{hypo:1}, the upscaling adjustment should be retained under this condition. 
For \(\eta_t,\frac{\eta_t}{\max(\lambda^t\beta, 1)} > \eta^*+e\) cases, we can derive that trial LR upscaling adjustment gives lower loss descent velocity. 
A downscaling adjustment should and will occur under such circumstance. 

\begin{proposition}
\label{prop:2}
    $\exists T' < \infty$, for $\forall t>T'$, the gap between $\eta_t$ and $\eta^*$ is bounded by the decaying LR scaling factors.
\begin{equation}
    \forall \eta_t, \frac{\eta^*-e}{\alpha'}\beta' < \eta_t < \frac{\eta^*+e}{\beta'}\alpha'
\end{equation}
where $\beta'=\frac{1}{\max(\lambda^t\beta, 1)}$ and $\alpha'=\max(\lambda^t\alpha, 1)$ are decaying LR adjustment factors. 
\end{proposition}

In practice, the value of $\lambda$ is set relatively large, ensuring that the learning rate $\eta$ converges to the neighborhood of $\eta^*$ at the current LR adjustment magnitude before the scaling factor becomes too small. 
Since the desired LR adjustment is guaranteed by Proposition~\ref{prop:1}, we begin with $\frac{\eta^*-e}{\alpha'}<\eta_t<\frac{\eta^*+e}{\beta'}$.
In this way, the learning rate $\eta_t$ either oscillates within this range, or attempts to search the learning rate in $[\frac{\eta^*+e}{\beta'},\frac{\eta^*+e}{\beta'}\alpha']$ and $[\frac{\eta^*-e}{\alpha'}\beta',\frac{\eta^*-e}{\alpha'}]$ respectively for higher or lower learning rate values. 
Since we have $\frac{\eta^*+e}{\beta'}>\eta^*+e$ always holds true, the LR upscaling attempts will trigger LR downscaling according to Proposition~\ref{prop:1}, maintaining the LR value given in Proposition~\ref{prop:2}. 
Similarly, learning rate downscaling attempts will be rejected by Proposition~\ref{prop:1}.

Proposition~\ref{prop:2} indicates that $\eta_t$ is bounded to the neighborhood of $\eta^*$ by decaying LR adjustment factors.
As the LR adjustment factor bound narrows, \textit{e.g.} $\mathop{\text{lim}}\limits_{t \rightarrow \infty} \frac{\alpha'}{\beta'} = 1$ and $\mathop{\text{lim}}\limits_{t \rightarrow \infty} \frac{\beta'}{\alpha'} = 1$, $\eta_t$ will fall into the $e$-neighborhood of $\eta^*$ eventually. 
Therefore, we have Theorem~\ref{theorem:1} proved. 





\subsection{Complexity Analysis}
We introduce the following theorem to demonstrate AdaLRS's geometric error decay. 
\begin{theorem}
\label{theorem:2}
    There exists $\gamma \in (0, 1)$ such that for $\eta_t \notin N_e(\eta^*)$, $|\eta_{t+k}-\eta^*| \leq \gamma|\eta_t-\eta^*|$ when LR adjustment is triggered.
\end{theorem}

For $\eta_t \notin N_e(\eta^*)$, let $\gamma_a \in (\frac{\eta^*-\alpha'\eta_0}{\eta^*-\eta_0},1)$ for the learning rate upscaling process. 
Since $\eta$ moves towards $\eta^*$ in the upscaling process, the following inequation holds for the whole upscaling process: 
\begin{equation}
    \forall t>0, \frac{\eta^*-\alpha'\eta_t}{\eta^*-\eta_t} < \frac{\eta^*-\alpha'\eta_0}{\eta^*-\eta_0}, |\eta_{t+k}-\eta^*| < \gamma|\eta_t-\eta^*|.
\end{equation}
Similarly, $\gamma_b$ can be selected from $(\frac{\beta'\eta_0-\eta^*}{\eta_0-\eta^*},1)$ for the LR downscaling process. 
Selecting $\gamma=max(\gamma_a,\gamma_b)$, Theorem~\ref{theorem:2} is proved directly. 

Theorem~\ref{theorem:2} implies the geometric error decay of the AdaLRS algorithm bounded by $\gamma$. 
Let $R$ be the range of the learning rate search space.
AdaLRS is able to approximate the optimal learning rate within $\mathcal{O}(\log R)$ adjustments. 
As a result, for foundation model pretraining tasks with data size larger than $\mathcal{O}(\log R)$, AdaLRS reaches the optimum neighborhood within a single run. 
The effectiveness of AdaLRS surpasses traditional auto hyperparameter search algorithms significantly, which often require extensive independent experiments to take effect. 


\vspace{-5pt}
\section{Experiments}
\vspace{-5pt}

\label{exp}

\subsection{Experiment Setup}

\textbf{Model Training. }
\label{sec: exp_setting_training}
We conduct experiments on both LLM and VLM pretraining. 
Qwen2.5-1.5B and Qwen2.5-7B~\cite{qwen2.5} are adopted for LLM pretraining, with parameters randomly initialized via He initialization~\cite{he2015delving}, Glorot initialization~\cite{glorot2010understanding}, and etc. 
We use a total of approximately 64M samples from the SlimPajama~\cite{cerebras2023slimpajama} dataset for LLM training from scratch, with all samples shuffled randomly. 
For VLM pretraining, on the other hand, we adopt the model structure of SAIL-VL~\cite{dong2025scalable}, with InternViT-300M~\cite{chen2023internvl} and Qwen2.5-1.5B adopted as backbone models. 
We use a collection of detail caption and image OCR data to train the vision-to-language projector from scratch. 
Detail caption datasets are curated via a similar recaption procedure as described in SAIL-VL~\cite{dong2025scalable}, while OCR data is collected from a series of opensource datasets~\cite{biten2022ocr,wang2020docstruct,Gupta16}. 

To demonstrate the effectiveness of the AdaLRS algorithm, we set different learning rates for each training task, \textit{i.e.}, learning rates too small, too large, and appropriate for pretraining.
For each setting, we train a baseline model and a model optimized by AdaLRS for fair comparison. 
We set the upscaling factor $\alpha$, downscaling factor $\beta$, and decaying factor $\lambda$ as $3$, $2$, and $0.99$ in all experiments for LR adjustment effectiveness. 
Approximately 120B and 160B tokens are used for LLM and VLM pretraining, with roughly 10,000 and 20,000 910B NPU hours consumed for 2B and 7B model pretraining experiments. 
A cosine learning rate scheduler~\cite{loshchilov2016sgdr} is applied in the main experiments. 
Detailed model training recipes are elaborated in Table~\ref{tbl: hyper}.

\begin{table*}[t]
\centering
\caption{
    Detailed hyperparameters for the main experiments. 
    ``Fit”, ``Large”, and ``Small” refer appropriate, too large, and too small learning rates, respectively.
    ``BSZ” stands for batch size. 
}
\resizebox{0.8\textwidth}{!}{
\begin{tabularx}{\textwidth}{l | X X X | X X X | X X X}
\toprule
\multicolumn{1}{c|}{} & \multicolumn{3}{c|}{\textit{2B LLM}} & \multicolumn{3}{c|}{\textit{7B LLM}} & \multicolumn{3}{c}{\textit{2B VLM}} \\
Hyperparameter & Fit & Large & Small & Fit & Large & Small & Fit & Large & Small\\
\midrule

Learning Rate & $2e^{-4}$ & $2e^{-3}$ & $2e^{-5}$ & $2e^{-4}$ & $2e^{-3}$ & $2e^{-5}$ & $8e^{-3}$ & $4e^{-1}$ & $2e^{-4}$  \\
BSZ / Micro BSZ & \multicolumn{3}{c|}{$1024/512$} & \multicolumn{3}{c|}{$2048/512$} & \multicolumn{3}{c}{$2048/1024$} \\
Window Size $k$ & \multicolumn{3}{c|}{$2500$} & \multicolumn{3}{c|}{$2000$} & \multicolumn{3}{c}{$1000$} \\
Data Composition & \multicolumn{3}{c|}{Detail Caption \& OCR} & \multicolumn{3}{c|}{SlimPajama Train Set} & \multicolumn{3}{c}{SlimPajama Train Set} \\
Search Step Ratio & \multicolumn{3}{c|}{$[0.1,0.4]$} & \multicolumn{3}{c|}{$[0.1,0.35]$} & \multicolumn{3}{c}{$[0.1,0.35]$} \\
\bottomrule
\end{tabularx}
}
\label{tbl: hyper}
\end{table*}

\textbf{Evaluation. }
To demonstrate the effectiveness of the proposed AdaLRS algorithm, we show the learning rate search and training loss dynamics of our foundation model training experiments. 
For LLM pretraining, we quantify the performance advantage of models trained with AdaLRS with final training loss and perplexity (PPL) computed on SlimPajama~\cite{cerebras2023slimpajama} train, validation, and test splits. 
We also conduct a lightweight SFT on 6M samples from the Infinity-Instruct dataset~\cite{li2025infinity}, and evaluate the downstream model performance on open-ended generation tasks such as Alpaca-Gen and KNIGHT-Gen~\cite{yang2025improving}. 
For VLM experiments, however, there is no widely accepted benchmark or metric for pretrained VLM evaluation.
Therefore, we tune the pretrained VLM on 3M samples from Infinity-MM~\cite{gu2024infinity} stage3 data with all parameters unfrozen, and use a series of publicly available VLM benchmarks for model evaluation, such as LLaVABench~\cite{liu2024visual}, MMVet~\cite{yu2024mm}, MMStar~\cite{chen2024we}, DocVQA~\cite{mathew2021docvqa}, OCRBench~\cite{liu2024ocrbench}, TextVQA~\cite{singh2019towards}, DetailCaps-4870~\cite{dong2024benchmarking}, etc.


\vspace{-3pt}
\subsection{Main Results}
\vspace{-3pt}

\begin{figure}[t]
    \centering
    \begin{subfigure}{0.31\textwidth}
        \includegraphics[width=\textwidth]{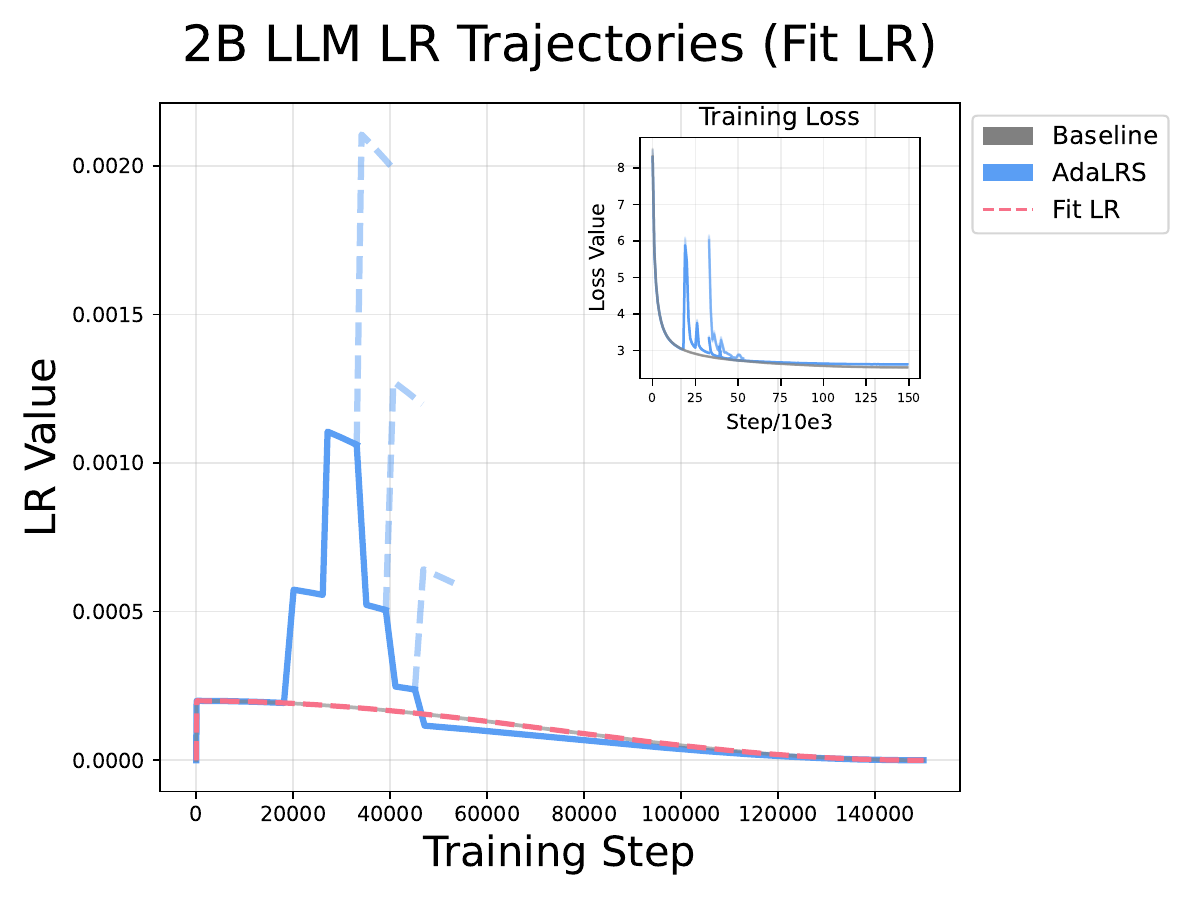}
        \caption{}
    \end{subfigure}
    \begin{subfigure}{0.31\textwidth}
        \includegraphics[width=\textwidth]{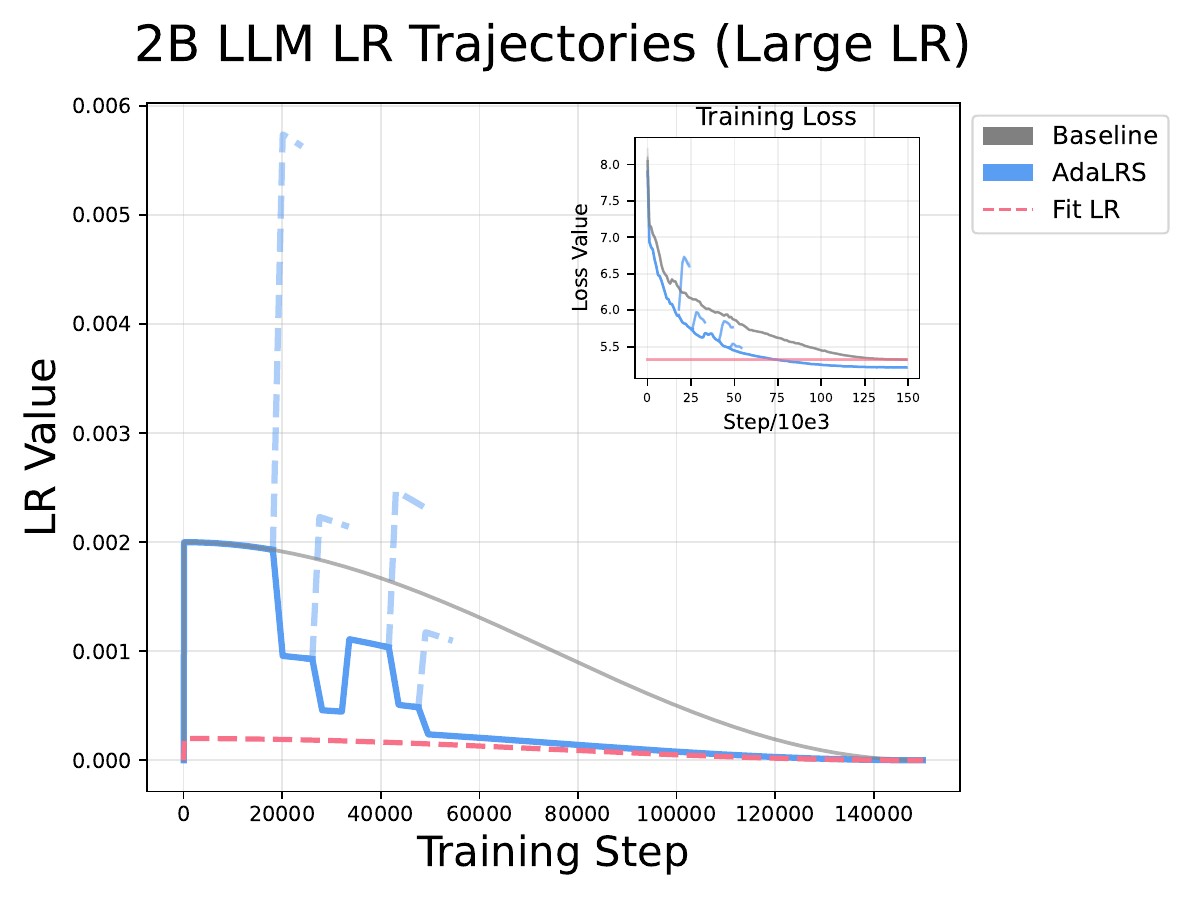}
        \caption{}
    \end{subfigure}
    \begin{subfigure}{0.31\textwidth}
        \includegraphics[width=\textwidth]{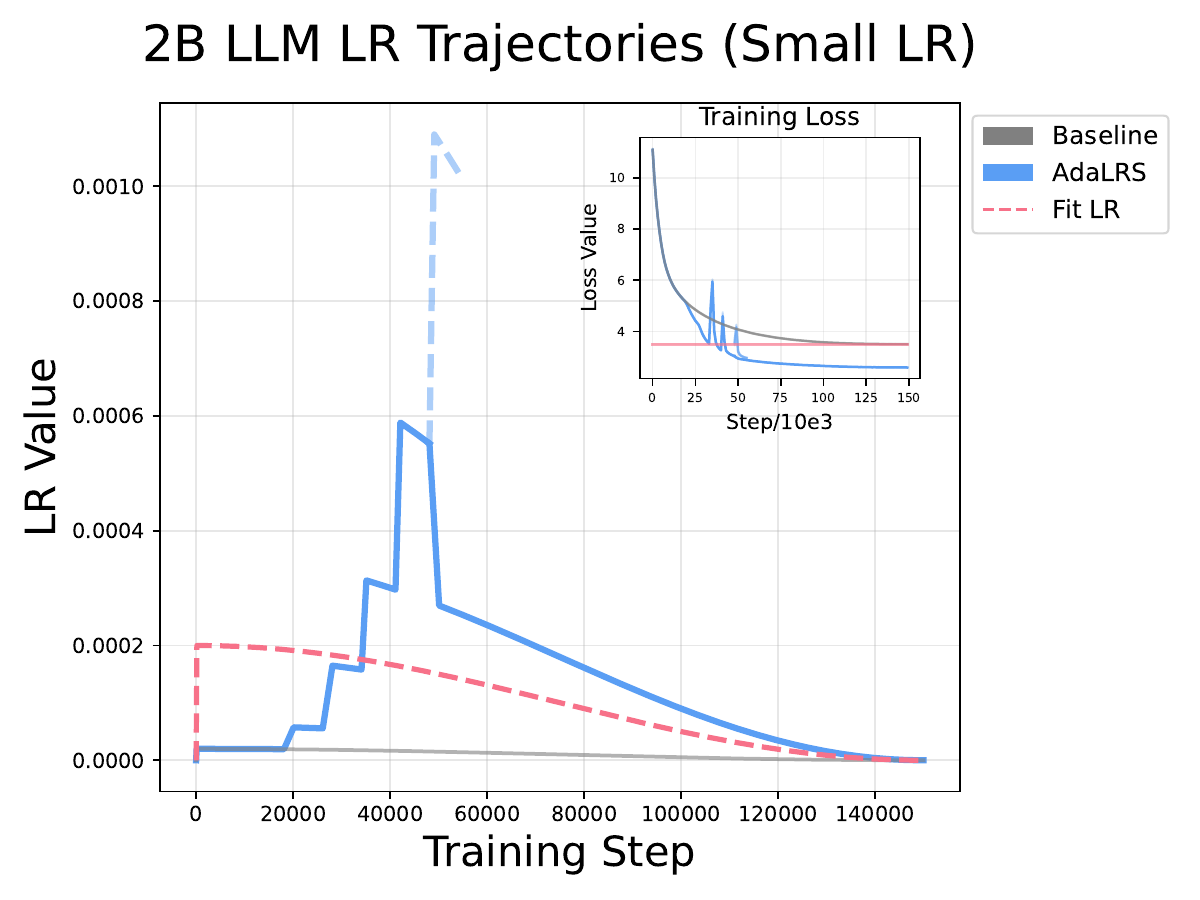}
        \caption{}
    \end{subfigure}

    \begin{subfigure}{0.31\textwidth}
        \includegraphics[width=\textwidth]{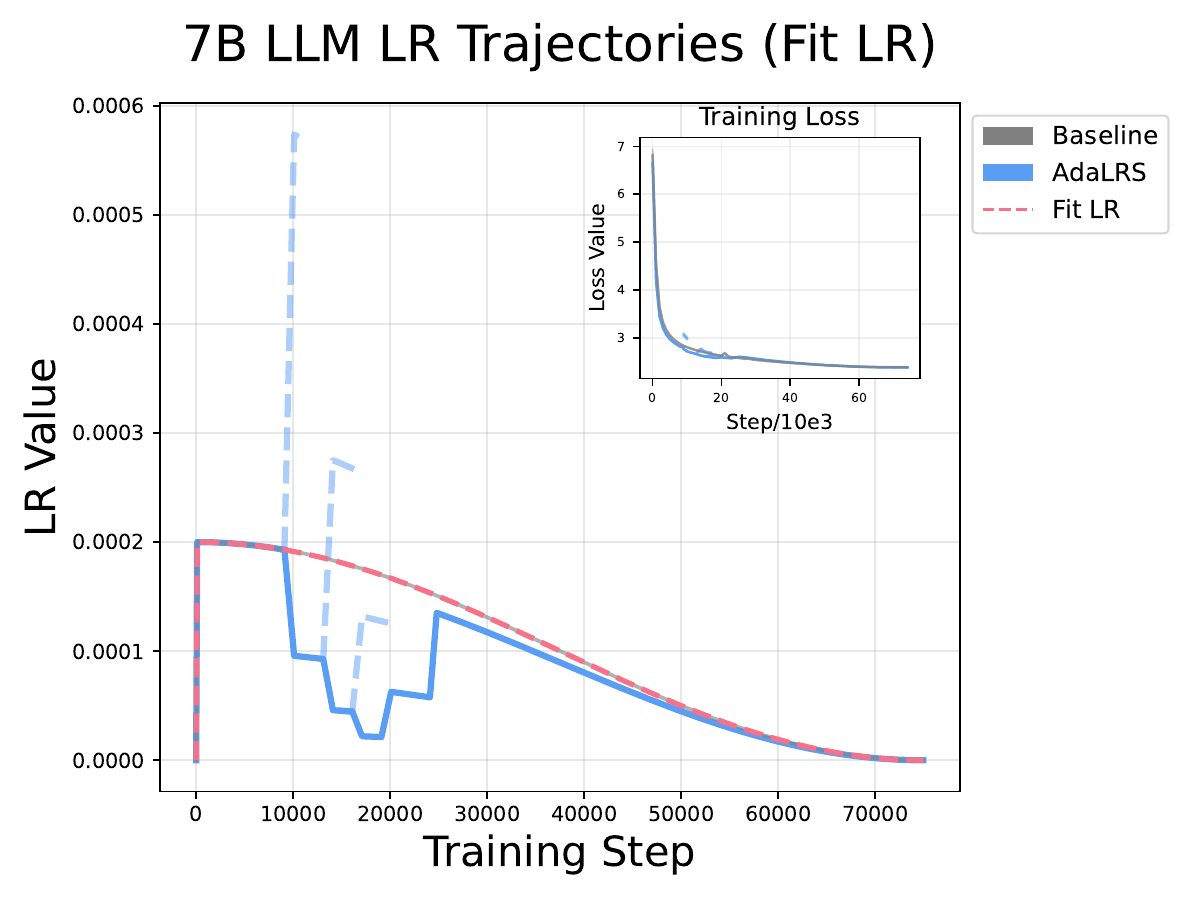}
        \caption{}
    \end{subfigure}
    \begin{subfigure}{0.31\textwidth}
        \includegraphics[width=\textwidth]{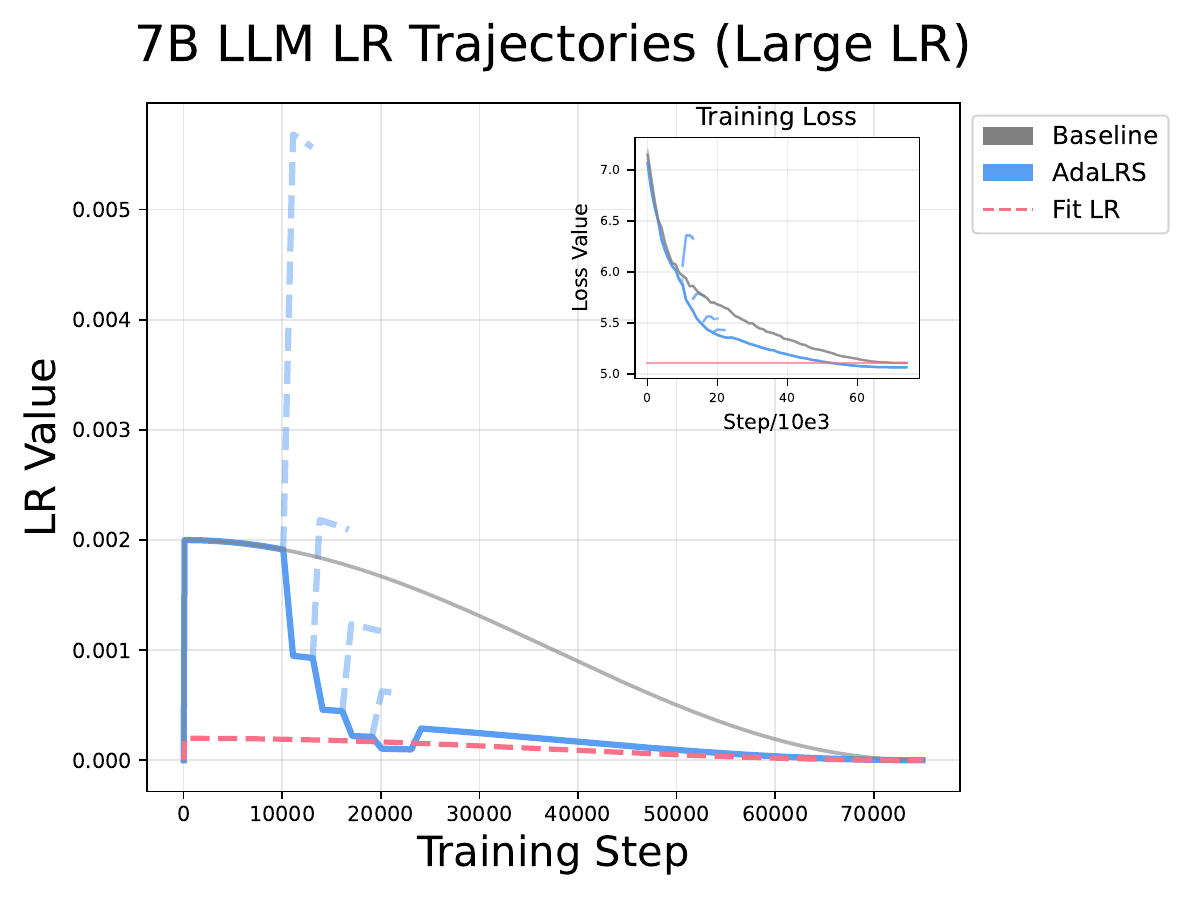}
        \caption{}
    \end{subfigure}
    \begin{subfigure}{0.31\textwidth}
        \includegraphics[width=\textwidth]{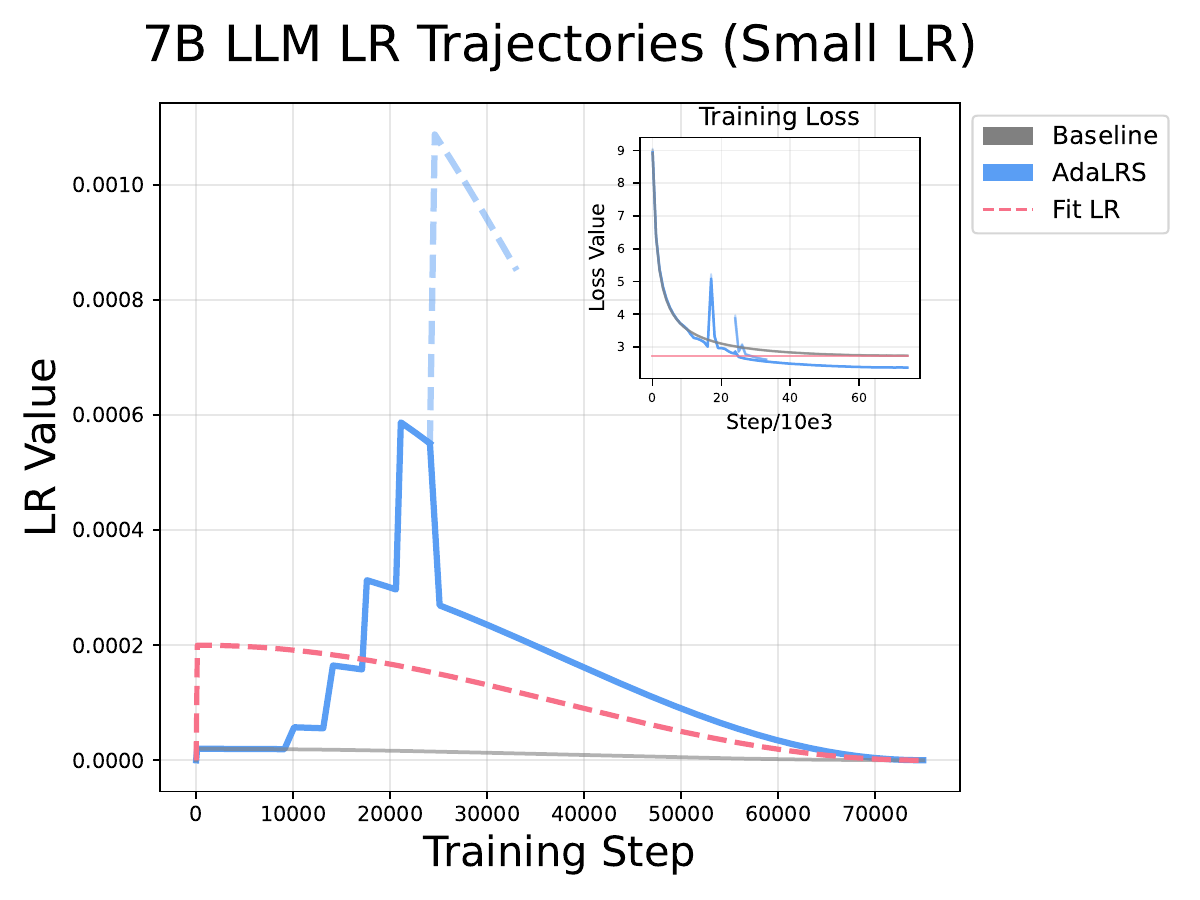}
        \caption{}
    \end{subfigure}

    \begin{subfigure}{0.31\textwidth}
        \includegraphics[width=\textwidth]{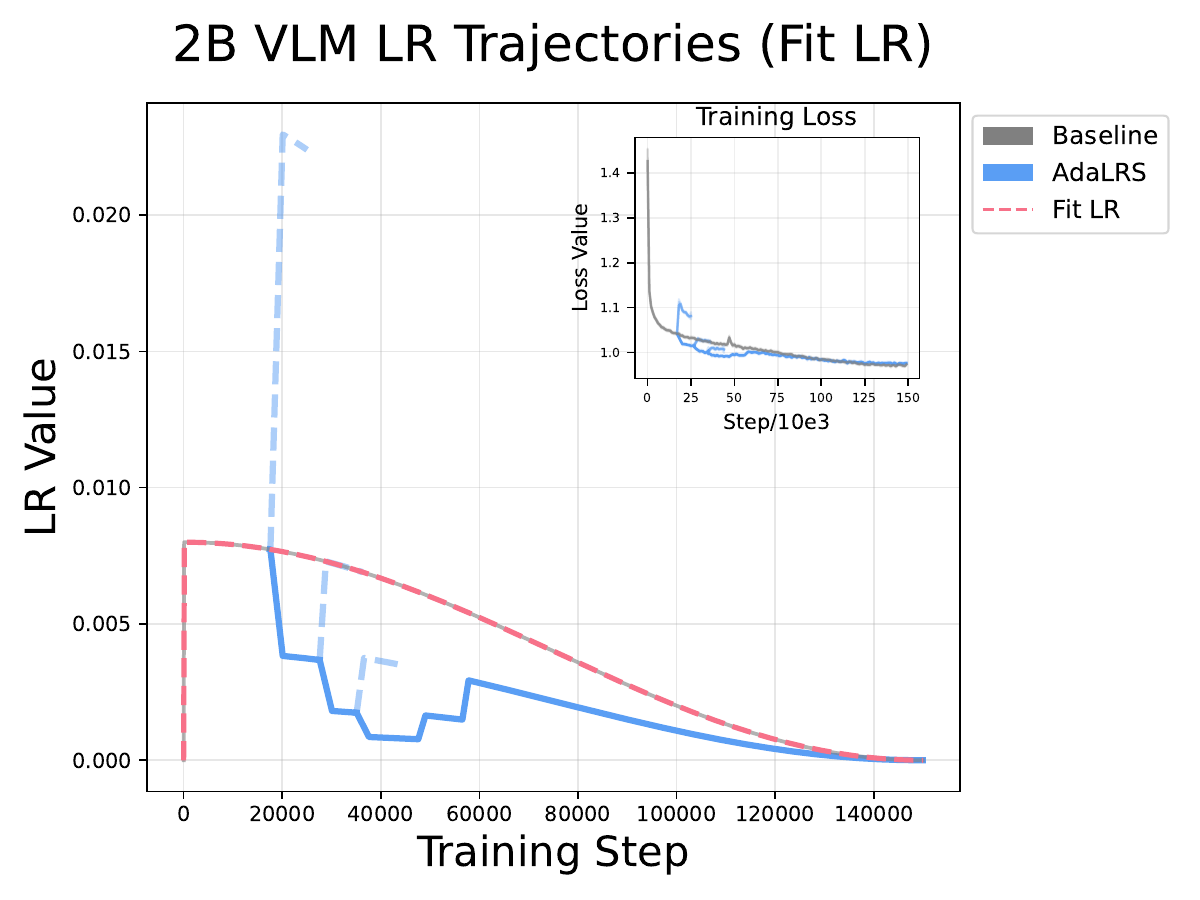}
        \caption{}
    \end{subfigure}
    \begin{subfigure}{0.31\textwidth}
        \includegraphics[width=\textwidth]{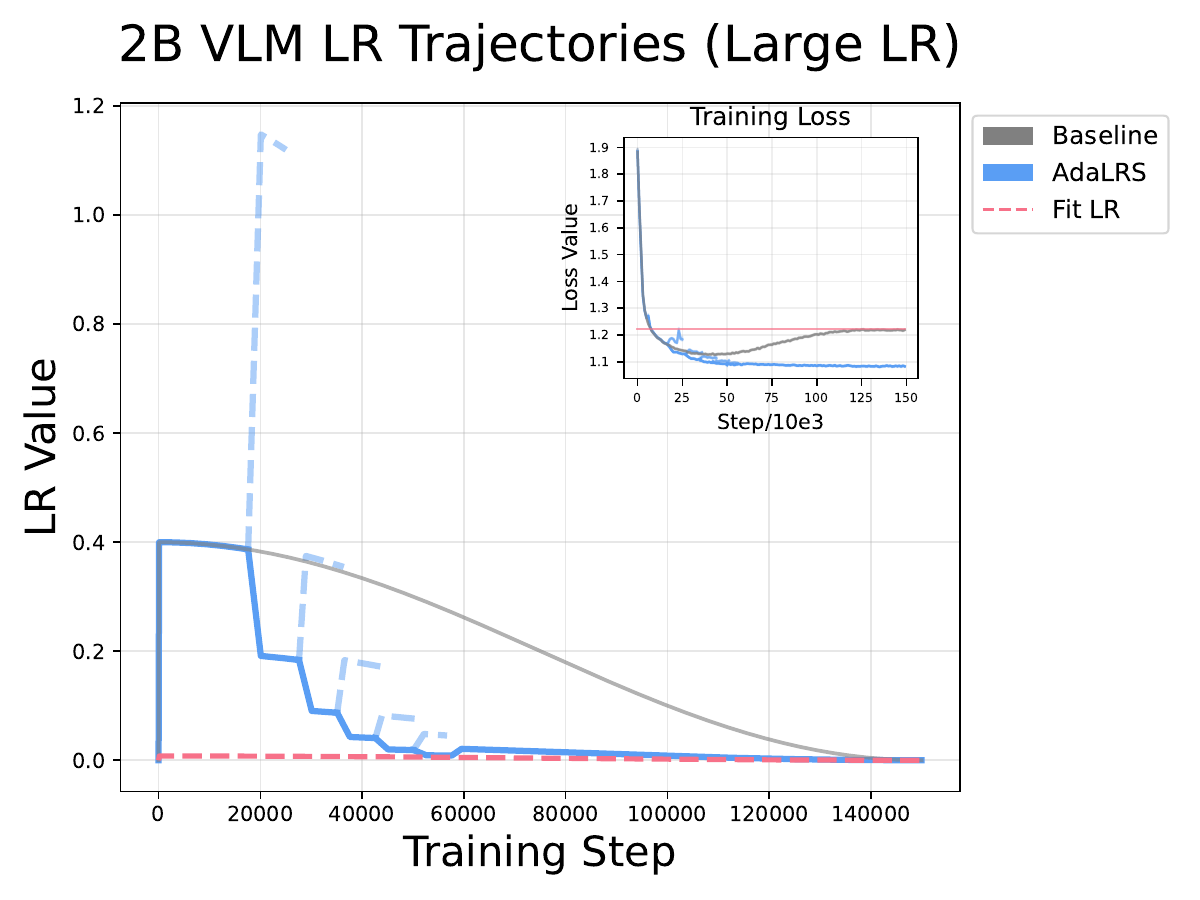}
        \caption{}
    \end{subfigure}
    \begin{subfigure}{0.31\textwidth}
        \includegraphics[width=\textwidth]{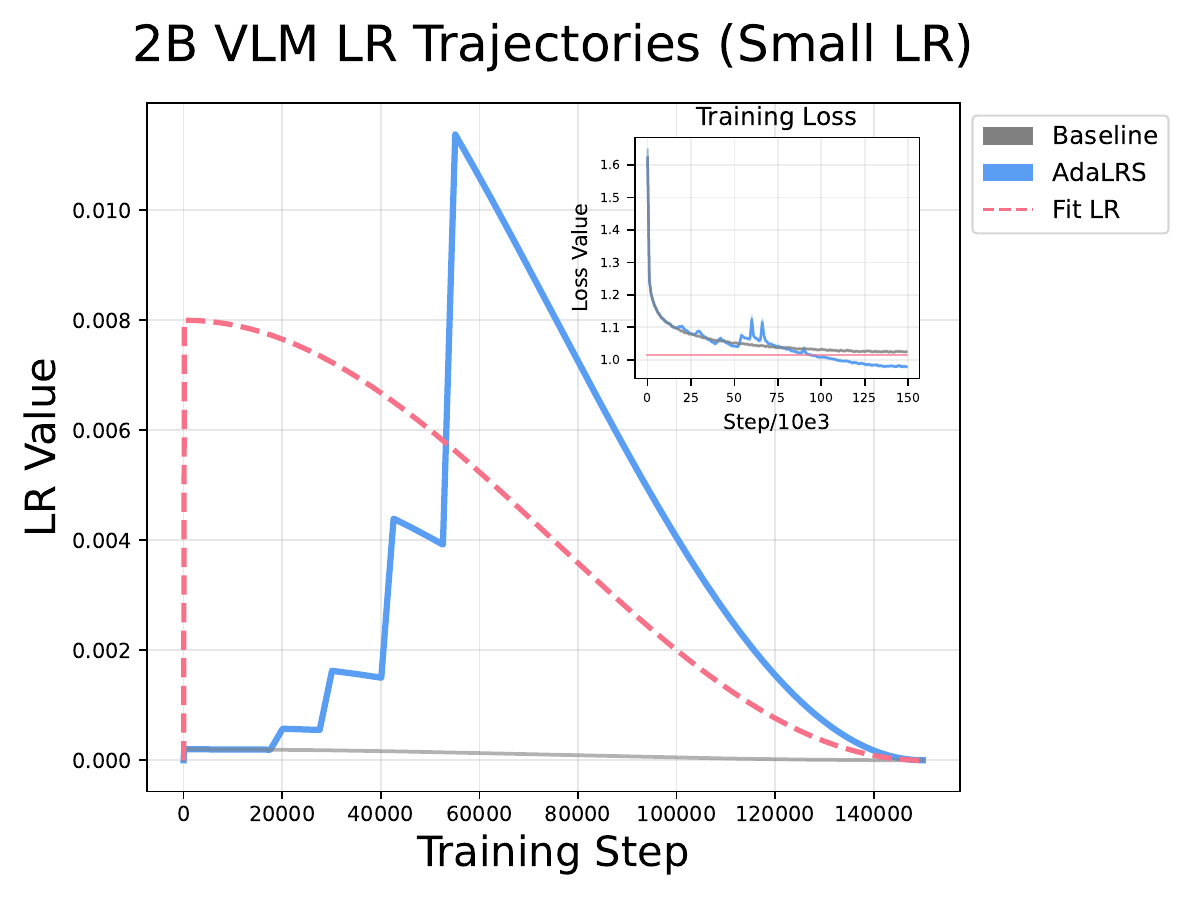}
        \caption{}
    \end{subfigure}
    
    \caption{
        AdaLRS's Learning rate adjustment process in foundation model pretraining under differnt LR settings.
        ``Fit LR” refers to learning rate appropriate for the pretraining task estimated by pilot study results. 
        Dashed curves represent failed LR upscaling attempts. 
    }
    \label{fig:main}
\end{figure}

\textbf{AdaLRS Approximates the Appropriate Learning Rate Effectively. }
As shown in Figure~\ref{fig:main}, the proposed AdaLRS algorithm (\textit{i.e.}, blue lines) pushes inappropriate learning rates (\textit{i.e.}, gray lines) towards the optimal ones (\textit{i.e.}, red dashed lines) effectively. 
For models trained with appropriate initial LR settings (Figures~\ref{fig:main}(a)(d)(g)), AdaLRS conducts trial attempts to adjust the learning rates and eventually converges to the vicinity of the optimum.
For excessively large or small LR settings (Figures~\ref{fig:main}(b)(c)(e)(f)(h)(i)), the adjusted learning rates produced by AdaLRS demonstrate marked advantages in consistency with the optimum compared with baseline experiments.

Although exhibiting compatibility with varying LR settings, several undesired LR scaling operations can be observed in Figures~\ref{fig:main}(c)(d)(f).
We attribute them to the aggressive upscaling and downscaling factor settings and the limited training steps for LR adjustment, which can be mitigated by milder LR scaling factors or longer LR search processes. 


\begin{table*}[t]
\caption{
    Evaluation results of our LLM pretraining experiments. 
    We use ``/” to separate performance scores of models trained with AdaLRS and baselines. 
    Black and gray numbers indicate better or worse performance. 
}
\resizebox{0.9\textwidth}{!}{
    \begin{tabular}{l | c c c | c c c}
    \toprule
     & \multicolumn{3}{c|}{\textit{\textit{2B LLM}}} & \multicolumn{3}{c}{\textit{\textit{7B LLM}}} \\
    Benchmark & \textbf{Fit LR} & \textbf{Large LR} & \textbf{Small LR} & \textbf{Fit LR} & \textbf{Large LR} & \textbf{Small LR} \\
    
    \midrule
    
    Training Loss        & \color{gray}$2.62/{\color{black}2.54}$ & $5.21/{\color{gray}5.32}$ & $2.56/{\color{gray}3.50}$ & $2.38/{\color{gray}2.39}$ & $5.07/{\color{gray}5.11}$ & $2.38/{\color{gray}2.74}$  \\
    Training PPL        & \color{gray}$13.72/{\color{black}12.65}$ & $183.28/{\color{gray}205.00}$ & $12.88/{\color{gray}33.21}$ & $10.84/{\color{gray}10.88}$ & $158.54/{\color{gray}165.67}$ & $10.76/{\color{gray}15.49}$  \\
    Validation PPL        & \color{gray}$13.51/{\color{black}12.42}$ & $183.94/{\color{gray}204.97}$ & $12.66/{\color{gray}32.69}$ & $10.67/{\color{gray}10.72}$ & $158.22/{\color{gray}165.70}$ & $10.61/{\color{gray}15.30}$  \\
    Test PPL        & \color{gray}$13.51/{\color{black}12.43}$ & $183.70/{\color{gray}204.68}$ & $12.66/{\color{gray}32.66}$ & $10.69/{\color{gray}10.74}$ & $158.16/{\color{gray}165.61}$ & $10.63/{\color{gray}15.32}$  \\

    \midrule
    
    Alpaca-Gen  & ${\color{gray}17.29}/18.56$ & $7.11/{\color{gray}6.09}$ & $17.10/{\color{gray}15.40}$ & $21.76/{\color{gray}21.61}$ & $6.35/{\color{gray}5.55}$ & $21.15/{\color{gray}20.68}$ \\
    KNIGHT-Gen  & ${\color{gray}10.13}/11.29$ & $3.96/{\color{gray}2.02}$ & $10.81/{\color{gray}8.36}$ & $13.62/{\color{gray}13.35}$ & $4.08/{\color{gray}3.82}$ & $13.53/{\color{gray}12.29}$ \\
    
    \bottomrule
    \end{tabular}
}
\centering
\label{tbl: llm_eval}
\vspace{-5pt}
\end{table*}

\begin{table*}[t]
\vspace{-3mm}
\caption{
    Evaluation results for the our 2B VLM experiments. 
    DetailCaps-4870 is used to evaluate the basic visual understanding ability of the pretrained model, while other benchmarks scores are evaluated with the SFT model. 
}
\resizebox{\textwidth}{!}{
    \begin{tabular}{l | c c c c c c c | c}
    \toprule
    LR Setting & \textbf{LLaVABench} & \textbf{MMVet} & \textbf{MMStar} & \textbf{DocVQA} & \textbf{OCRBench} & \textbf{TextVQA} & \textbf{DetailCaps-4870} & \textbf{Average} \\
    
    \midrule

    Fit LR        & $39.5/{\color{gray}38.5}$ & $34.58/{\color{gray}32.02}$ & $\color{gray}48.67/{\color{black}49.53}$ & $\color{gray}77.99/{\color{black}78.00}$ & $\color{gray}718/{\color{black}735}$ & $64.89/{\color{gray}63.74}$ & $55.68/{\color{gray}55.30}$ & $56.16/{\color{gray}55.80}$  \\
    Large LR        & $36.8/{\color{gray}35.7}$ & $31.47/{\color{gray}30.50}$ & $44.47/{\color{gray}44.33}$ & $57.53/{\color{gray}57.42}$ & $631/{\color{gray}606}$ & $60.32/{\color{gray}58.08}$ & $49.02/{\color{gray}47.08}$ & $48.96/{\color{gray}47.67}$  \\
    Small LR        & $44.3/{\color{gray}39.2}$ & $36.15/{\color{gray}30.23}$ & $\color{gray}48.67/{\color{black}49.20}$ & $77.75/{\color{gray}77.47}$ & $730/{\color{gray}689}$ & $64.85/{\color{gray}57.86}$ & $56.65/{\color{gray}53.51}$ & $57.34/{\color{gray}53.77}$  \\

    \bottomrule
    \end{tabular}
}
\centering
\label{tbl: vlm_eval}
\vspace{-3pt}
\end{table*}

\textbf{AdaLRS Accelerates Pretraining Convergence Significantly. }
\label{sec:main_res_2}
As shown in the top-right corner of each subfigure in Figure~\ref{fig:main}, models trained with AdaLRS demonstrate significant advantages in training losses. 
For learning rates too small or too large, models trained with vanilla cosine LRS suffer from slow loss descent velocities, with severe training instability observed under certain circumstances (Figure~\ref{fig:main}(h)).
AdaLRS not only improves training loss convergence effectively, but also aids such training instability problems. 
In experiments with appropriate learning rate settings, AdaLRS introduces slight loss spikes in model training, but the resulted training loss remains close to the baseline, exhibiting desired stability. 

We also illustrate the acceleration ratio of our method for inappropriate LR settings.
As shown in the training loss curves of Figures~\ref{fig:main}(b)(c)(e)(f)(h)(i), AdaLRS experiments reach the final training loss of the baselines at early training stages. 
For excessively small learning rates, AdaLRS surpasses the baselines at approximately $50\%$ training steps, while more than $30\%$ training costs can be reduced for large LR settings. 
Although the LR upscaling attempts may introduce extra training costs, significant training loss advantages are still witnessed in models trained with AdaLRS under exactly the same training budgets. 

It is also worth noticing that although AdaLRS adjusts learning rates reasonably for large LR settings, the training loss remains relatively high throughout the training process. 
We attribute this problem to the disruption of model initialization parameters due to the overly large learning rates. 
However, AdaLRS still reaches the vicinity of the optimal learning rate in a single run under such circumstances, which still improves the efficiency of the learning rate search process. 

\textbf{AdaLRS Enhances Pretrained Foundation Model Performance Markedly. }
To demonstrate the model performance advantage of training with AdaLRS, we provide PPL and downstream task evaluation results for LLM experiments in Table~\ref{tbl: llm_eval}. 
AdaLRS not only surpasses baseline LLMs pretrained with suboptimal learning rates, but also achieves comparable model performance with Fit LR baselines. 
It is worth noticing that AdaLRS improves model performance significantly for excessively small LRs, even surpassing the Fit LR baselines in 7B LLM pretraining. 
This observation implies that starting with a relatively small LR setting, AdaLRS can potentially achieve both the optimal learning rate and model performance for unexplored pretraining tasks within a single run. 

For VLM pretraining, we show VQA benchmark scores of the SFT VLM models in Table~\ref{tbl: vlm_eval}.
Models pretrained with the AdaLRS algorithm demonstrate remarkable performance advantages across most of the benchmarks, which encompass visual understanding tasks in both natural scenes and OCR-related tasks. 
Similar to experiments on LLM pretraining, we also observe that AdaLRS starting from a small learning rate yields better performance than Fit LR baselines, which validates the effectiveness of the AdaLRS algorithm in unexplored pretraining tasks.



\subsection{Extending AdaLRS to WSD Scheduler}
\begin{figure}[t]
    \centering
    \begin{subfigure}{0.31\textwidth}
        \includegraphics[width=\textwidth]{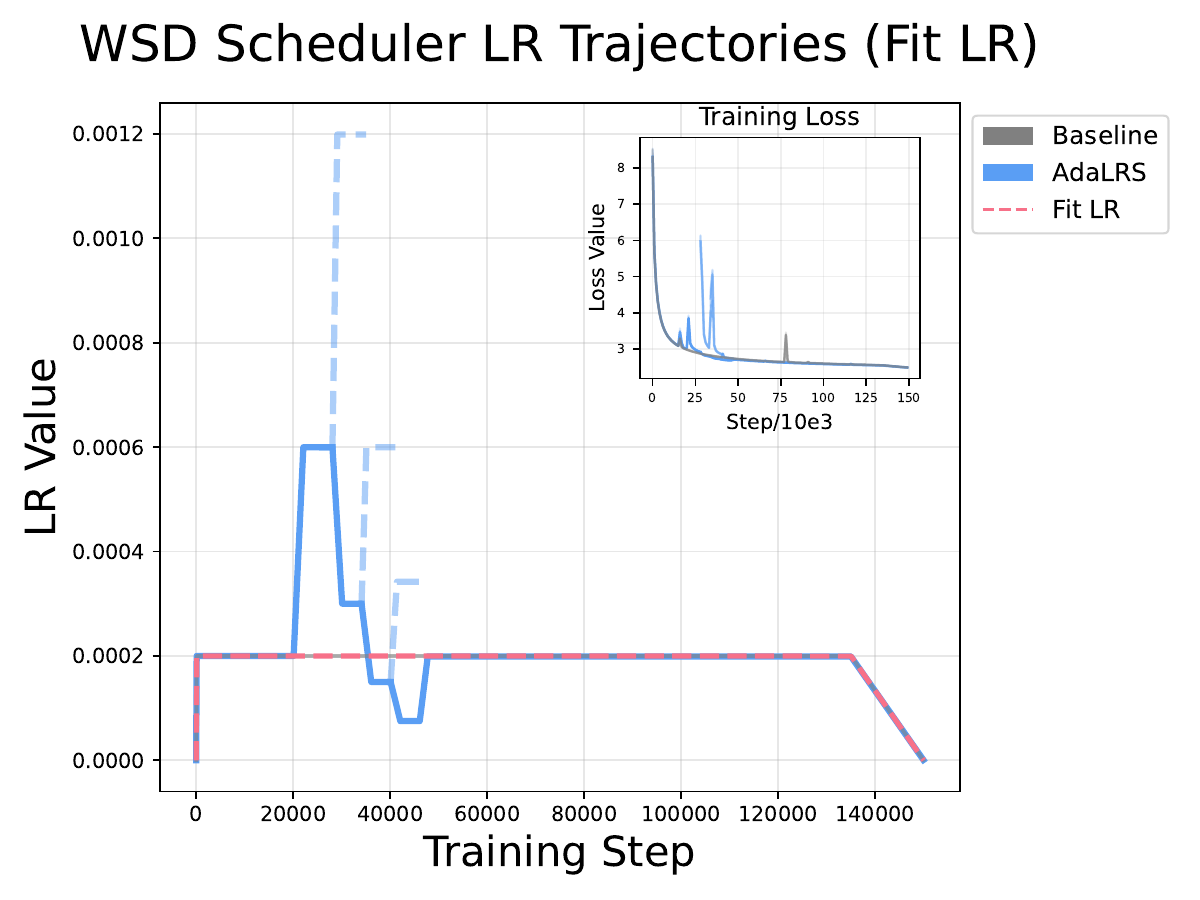}
    \end{subfigure}
    \begin{subfigure}{0.31\textwidth}
        \includegraphics[width=\textwidth]{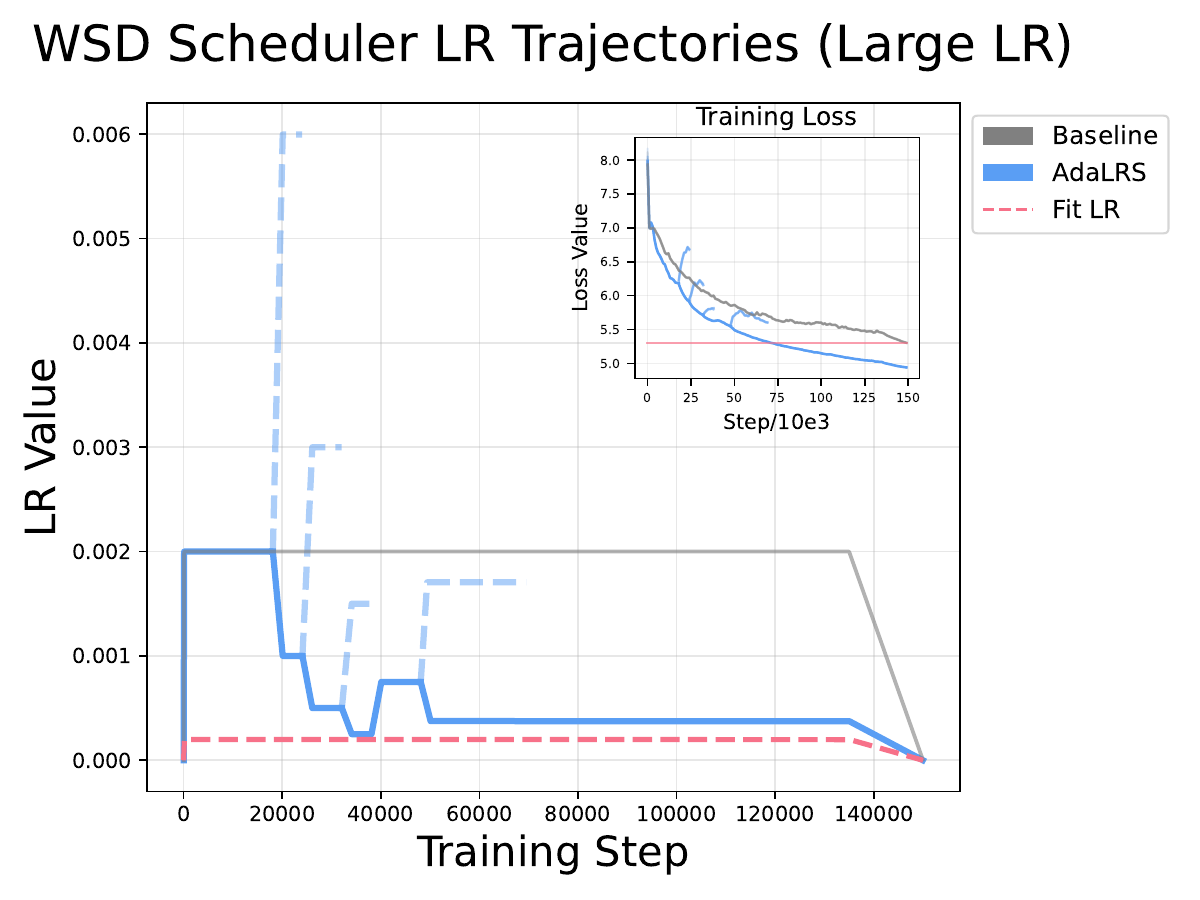}
    \end{subfigure}
    \begin{subfigure}{0.31\textwidth}
        \includegraphics[width=\textwidth]{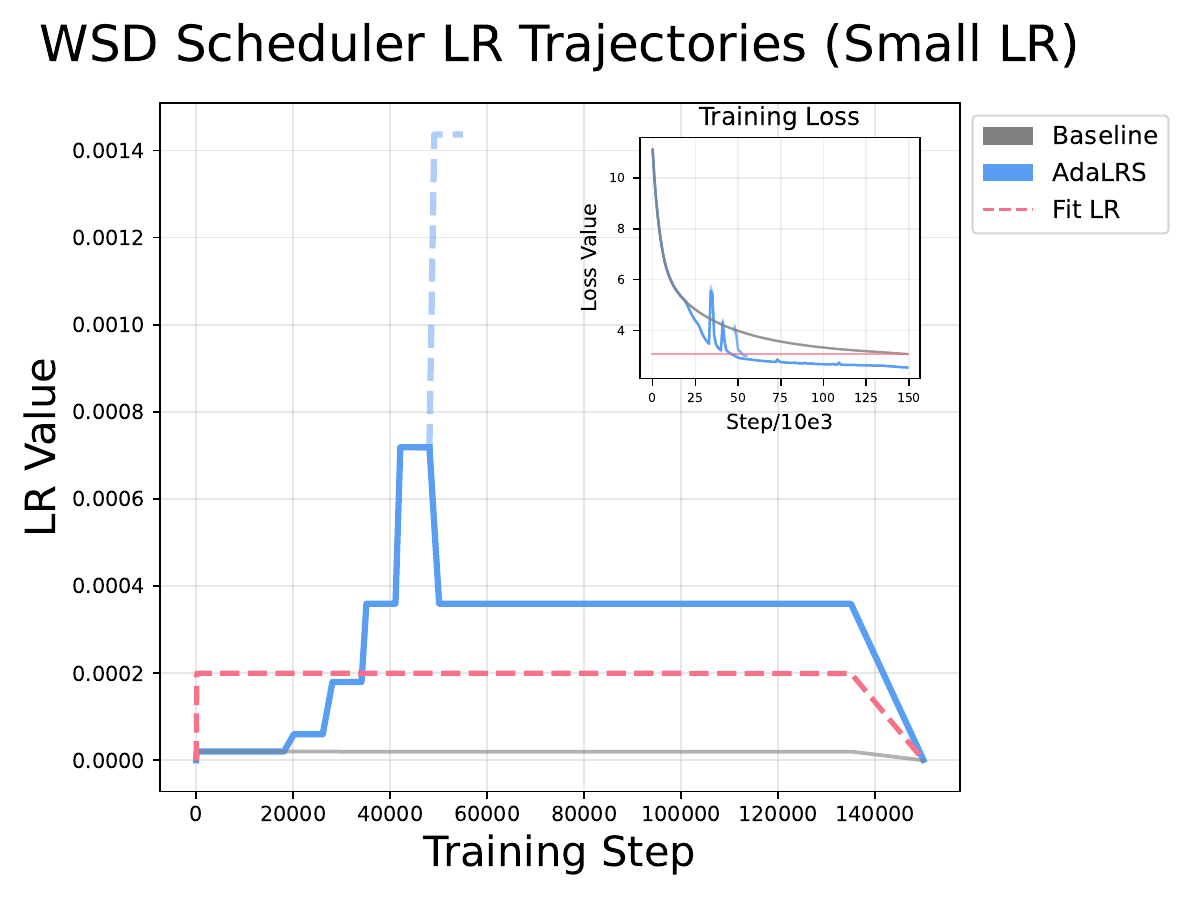}
    \end{subfigure}
    
    \caption{
        AdaLRS's Learning rate adjustment process in 2B LLM pretraining with WSD scheduler.
        We refer to Figure~\ref{fig:main} for denotation definitions. 
    }
    \label{fig:wsd}
    \vspace{-5pt}
\end{figure}

Since AdaLRS simply applies scaling factors on the learning rates produced by the scheduler and conducts LR search only in the early training stages, we can apply it to many mainstream schedulers seamlessly. 
Considering the generalizability of the conclusion and experiment efficiency, we use WSD~\cite{hu2024minicpm} scheduler instead of Cosine scheduler in our experiments, to demonstrate the compatibility of AdaLRS with other schedulers. 
We use a linear decay function for the final $10\%$ training steps in the implementation of the WSD scheduler. 
Experiment results are shown in Figure~\ref{fig:wsd}.
Similar to the experiment results on cosine learning rate schedulers shown in Figure~\ref{fig:main}(a)(b)(c), AdaLRS adjusts unreasonable LR settings effectively, and shows desired stability for appropriate configurations.

\subsection{Robustness Across Hyperparameter Settings}
In this part, we investigate the model performance dynamics with different AdaLRS hyperparameters. 
We take 2B LLM pretraining as an example and conduct experiments with small and large learning rate settings, respectively. 
As shown in Table~\ref{tbl: hyperparam_abl}, models trained with AdaLRS demonstrate marked advantages in training loss and evaluation PPL compared to baselines, showing strong robustness across varying hyperparameter combinations. 
We also observe a clear trend that, as the LR scaling and decay factors decrease, the final learning rate adjusted by AdaLRS tends to decrease for small-LR settings and increase for large-LR settings, respectively. 
In the experiments with $\alpha,\beta,\lambda=2,1.67,0.99$ and $\alpha,\beta,\lambda=2,1.67,0.95$, both settings yield nearly the same final learning rate, despite their different hyperparameters. This is because LR overshooting in the former triggers a downscaling adjustment, whereas the $0.95$ decay factor in the latter prevents such adjustment.

\begin{table*}[t]
\vspace{-3mm}
\caption{
    Model performance of 2B LLMs trained with different AdaLRS hyperparameters. 
    $\alpha$, $\beta$, and $\lambda$ refer to LR upscaling factor, downscaling factor, and decay factor, respectively. 
    Baselines with AdaLRS disabled are shown with ``$-$” as hyperparameters. 
}
\resizebox{0.9\textwidth}{!}{
    \begin{tabular}{l | c c c c c c | c c c c c}
    \toprule

    & \multicolumn{6}{c|}{\textbf{Small LR}} & \multicolumn{5}{c}{\textbf{Large LR}} \\
    $\alpha/\beta$ & $-$ & $3/2$ & $2/1.67$ & $1.5/1.43$ & $2/1.67$ & $2/1.67$ & $-$ & $3/2$ & $2/1.67$ & $1.5/1.43$ & $2/1.67$ \\
    $\lambda$ & $-$ & $0.99$ & $0.99$ & $0.99$ & $0.95$ & $0.9$  & $-$ & $0.99$ & $0.99$ & $0.99$ & $0.9$ \\
     
    \midrule
    Final LR        & $2.0e^{-5}$ & $3.6e^{-4}$ & $3.1e^{-4}$ & $1.5e^{-4}$ & $3.1e^{-4}$ & $2.2e^{-4}$ & $2.0e^{-3}$ & $3.8e^{-4}$ & $5.2e^{-4}$ & $7.2e^{-4}$ & $7.9e^{-4}$ \\
    Training Loss        & $3.07$ & $2.55$ & $2.55$ & $2.58$ & $2.54$ & $2.54$ & $5.30$ & $4.94$ & $5.10$ & $5.15$ & $5.08$  \\
    Validation PPL        & $21.59$ & $12.87$ & $12.75$ & $13.11$ & $12.61$ & $12.68$ & $201.54$ & $140.13$ & $163.95$ & $172.01$ & $159.55$  \\
    Test PPL        & $21.61$ & $12.89$ & $12.77$ & $13.13$ & $12.63$ & $12.70$ & $201.36$ & $140.07$ & $163.81$ & $171.85$ & $159.47$  \\

    \bottomrule
    \end{tabular}
}
\centering
\label{tbl: hyperparam_abl}
\vspace{-3pt}
\end{table*}

\subsection{Backtracking LR Downscaling Strategy Ablation}

\begin{figure}[t]
    \centering
    \begin{subfigure}{0.31\textwidth}
        \includegraphics[width=\textwidth]{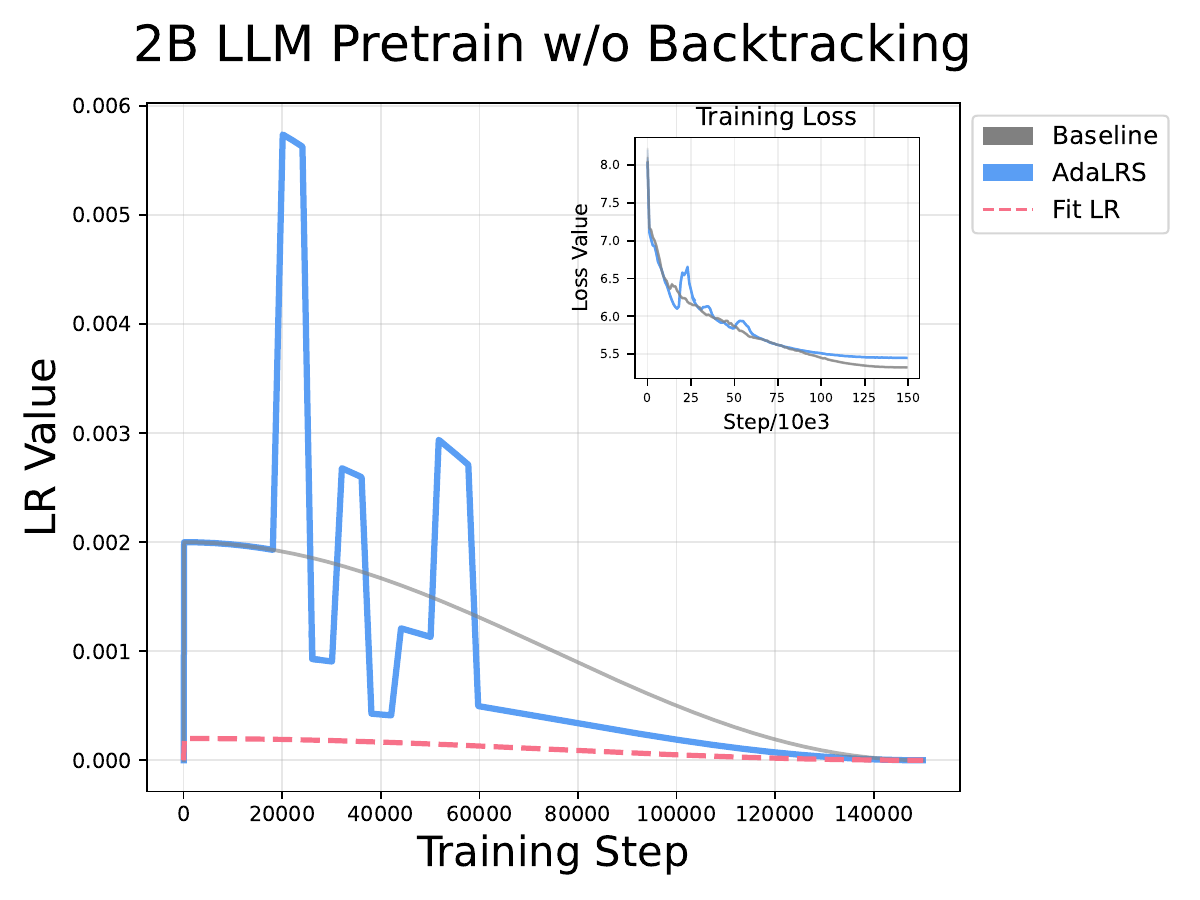}
        \caption{}
    \end{subfigure}
    \begin{subfigure}{0.31\textwidth}
        \includegraphics[width=\textwidth]{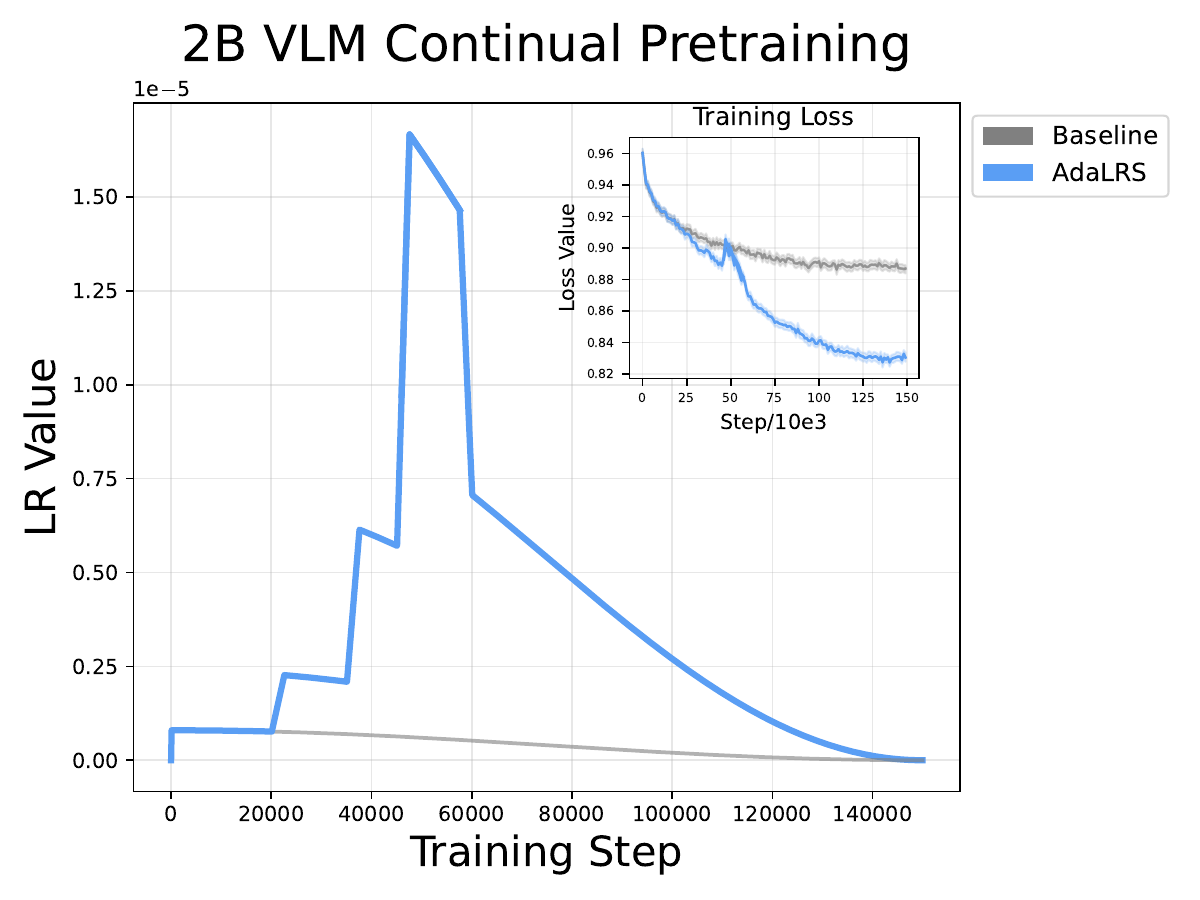}
        \caption{}
    \end{subfigure}
    \begin{subfigure}{0.31\textwidth}
        \includegraphics[width=\textwidth]{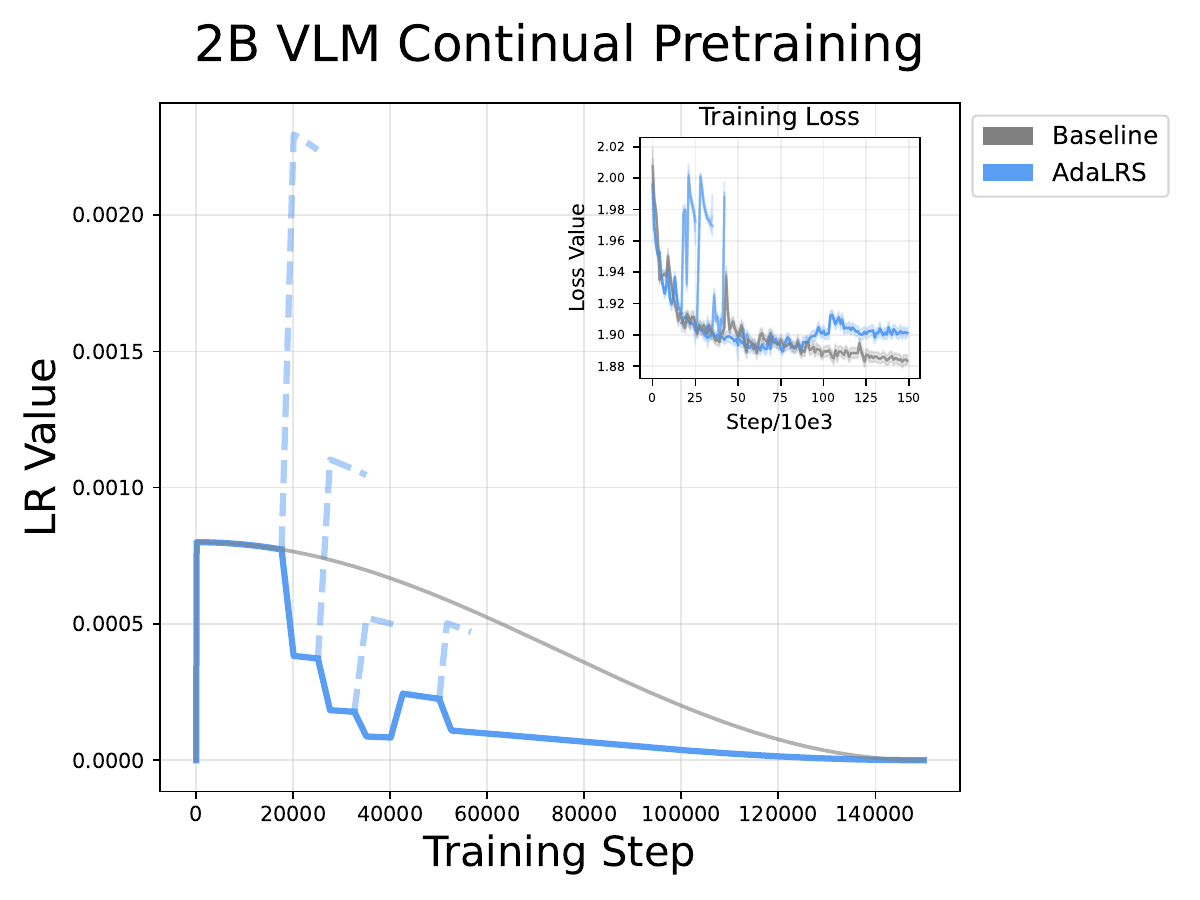}
        \caption{}
    \end{subfigure}

    \vspace{-1pt}

    \caption{
        Ablation studies for the backtracking downscaling strategy (a) and the training dynamics of AdaLRS on VLM continual pretraining (b)(c).
    }
    \label{fig:bt_and_ct}
    \vspace{-5pt}
\end{figure}

In this part, we evaluate the effectiveness of our backtracking LR downscaling strategy, which restores model training states before LR downscaling for training stability. 
We show model training dynamics w/o the backtracking strategy applied on the 2B LLM pretraining task in Figure~\ref{fig:bt_and_ct}, and the scaling factors are set the same as mentioned in Section~\ref{sec: exp_setting_training}.

As shown in Figure~\ref{fig:bt_and_ct}(a), the training loss curve moves upwards during LR upscaling attempts, which may introduce instability in model parameter distribution. 
Although AdaLRS adjusts the learning rate downwards to approximately $38\%$ of the baseline, the training loss remains relatively high in the entire training process. 
At the end of training, AdaLRS algorithm without backtracking yields higher training loss than the baseline. 
We refer to Figure~\ref{fig:main}(b) for training loss dynamics with the backtracking strategy applied. 
Loss upward movements and underlying disruptive parameter updates are eliminated by restoring model states. 
Comparing Figure~\ref{fig:bt_and_ct}(a) and Figure~\ref{fig:main}(b), the backtracking strategy improves the resulting model training loss by a large margin.

\subsection{AdaLRS for Continual Pretraining}
\label{sec:exp_ct}
In this part, we explore the compatibility of the proposed AdaLRS algorithm with the continual pretraining paradigm. 
Different from pretraining foundation models from scratch, disruptive parameter updates may result in catastrophic forgetting and harm model performance markedly in continual pretraining~\cite{luo2023empirical}. 
We take VLM continual pretraining as an example, updating both the projector and vision encoder simultaneously to enable more model capacity used for visual understanding.
It is a common practice to improve VLM model performance~\cite{bai2023qwen,yao2024minicpm,chen2024expanding}.
To be specific, we conduct 2B VLM continual pretraining from the model checkpoint after from-scratch pretraining, \textit{i.e.}, the baseline model in Figure~\ref{fig:main}(g). 
We locate the optimal learning rate in this stage at approximately $4e^{-5}$, with $8e^{-7}$ and $8e^{-4}$ used as small and large LR, respectively. 
The same set of VLM pretraining datasets is used in this stage, and the experiment results are shown in Figure~\ref{fig:bt_and_ct}(b)(c). 

As expected, AdaLRS adjusts learning rates towards the optimum effectively in VLM continual pretraining. 
For the small learning rate experiment shown in Figure~\ref{fig:bt_and_ct}(b), AdaLRS improves the training loss by a large margin from $0.8851$ to $0.8286$.  
However, although AdaLRS downscales the large learning rate in Figure~\ref{fig:bt_and_ct} (c), the final training loss remains higher than $1.88$. 
The LR downscaling adjustments fail to improve model performance and even introduce severer loss oscillation, resulting in higher training loss. 
These observations show that AdaLRS performs promisingly for continual foundation model training with small initial learning rates, but can not eliminate the catastrophic forgetting problem caused by large learning rates.

\vspace{-5pt}
\section{Related Work}
\vspace{-5pt}

\textbf{Optimal Learning Rate Prediction in Foundation Model Pretraining.}
Due to the prohibitively expensive pretraining costs, a large number of research seeks to predict the optimal learning rate for large scale LLM pretraining. 
This line of work typically focuses on summarizing model performance dynamics as a function \textit{w.r.t.} hyperparameter settings~\cite{kaplan2020scaling,bi2024deepseek,bjorck2024scaling,li2025predictable}, such as model size, batch size, training budget, etc.
MM1~\cite{mckinzie2024mm1} further extends this approach to VLM pretraining, discovering the dependency of optimal LR on the number of model parameters.
Although reducing hyperparameter search costs for larger scale model training, these methods require hundreds or even thousands of pretraining experiments on smaller models to form an optimal learning rate expression.
What's worse, the resulted optimal LR expressions are often restricted by certain model structure or data composition designs to take effect, suffering from poor generalizability.  


\textbf{Transferring Hyperparameter across Model Sizes. }
Tensor Program series work studies the transferability of hyperparameter settings across varying model sizes. 
Recently proposed $\mu$P~\cite{yang2021tensor} and $\mu$Transfer~\cite{yang2022tensor} prove the transferability of hyperparameter settings (including learning rate) across Multi-Layer Perceptron~\cite{popescu2009multilayer} (MLP) and Transformers~\cite{vaswani2017attention,devlin2018bert,brown2020language}. 
Powered by $\mu$Transfer, optimal learning rate search can be performed at proxy models and then transferred to larger ones~\cite{dey2023cerebras,hu2024minicpm}, reducing the search cost by a large margin. 
However, searching for the optimal learning rate on proxy models can still be resource-consuming, and the optimal learning rate may shift across model sizes in complicated model designs.

\textbf{Auto Hyperparameter Search. }
This line of work treats hyperparameter search as an optimization problem.
Since learning rate is a non-differentiable variable in model training, researchers propose to search for optimal learning rates via grid/random search~\cite{bergstra2012random}, Bayesian optimization~\cite{snoek2012practical}, multi-armed bandit algorithm~\cite{li2018hyperband}, evolution-based method~\cite{jaderberg2017population}, reinforcement learning~\cite{mounjid2024improving}, etc.
Nevertheless, despite the search cost reduction compared with brutal search, these methods still require extensive independent runs to establish the underlying optimal learning rate prediction model.

It is worth noticing that a recent work Hypergradient Descent (HD)~\cite{baydin2017online} also conducts online LR adjustments. 
However, HD serves as a learning rate scheduler in model training and is only verified to be effective against constant schedulers in small neural networks.
AdaLRS differs from HD in that our method is a LR search algorithm which is compatible with modern base LRSs, such as cosine and WSD schedulers, and is examined to be effective for modern foundation model pretraining.



\vspace{-5pt}
\section{Conclusions}
\vspace{-5pt}

In this paper, we propose AdaLRS, an online optimal learning rate search algorithm which optimizes the loss descent velocity to approximate the optimal learning rate. 
To validate this algorithm, we provide both theoretical and experimental analyzes to show the convexity of foundation model pretraining loss and its slope, as well as their shared optimum in LR. 
We also provide convergence and complexity analysis of AdaLRS to show its effectiveness and efficiency. 
Experiments in a series of foundation model pretraining tasks demonstrates that AdaLRS effectively adjusts inappropriate learning rates to the vicinity of the optimum in a single run and accelerates model convergence speed by a large margin. 
Starting from excessively small learning rates, AdaLRS achieves comparable and even superior performance with baseline models trained with near-optimal learning rates. 
AdaLRS is validated to improve model performance with varying model sizes, training paradigms, and base schedulers, exhibiting promising potential to be applied in unexplored pretraining tasks. 
We refer to Appendix~\ref{sec:limit} for the discussion of the limitations.

\newpage
\bibliographystyle{plain} 
\bibliography{reference}

\newpage
\clearpage
\appendix

\newpage

\section{AdaLRS Algorithm}
\label{appendix:adalrs_algo}
In this section, we show the workflow of the proposed AdaLRS algorithm in detail.
As shown in Algorithm~\ref{alg:adalrs}, AdaLRS tracks the velocity of loss descent in pretraining tasks and computes the velocity as the loss curve slope via the least squares method~\cite{bjorck1990least}. 
Learning rate upscaling is triggered when the velocity decays, followed by several validation steps to measure the loss descent velocity after LR adjustment.
We then compare the loss descent velocity after LR adjustment with that corresponding to similar loss values before upscaling. 
If the velocity becomes larger, the upscaling adjustment is retained; otherwise, the model and optimizer states are restored to the exact step before LR upscaling, followed by a LR downscaling adjustment instead. 
Restoring training states is designed to stabilize the training process, where the upscaled learning rate may disrupt model parameter distribution during training. 
After several loops of learning rate adjustment, the resulting learning rate eventually falls in the neighborhood of the optimum.

\begin{algorithm}[H]
\caption{Adaptive Learning Rate Scheduling (AdaLRS)}
\label{alg:adalrs}
\DontPrintSemicolon
\KwIn{
    Initial learning rate $\eta_0$, Upscale factor $\alpha > 1$, Downscale factor $\beta> 1$ ($\gcd(\alpha, \beta) = 1$), Scale decay factor $\lambda<1$, Loss slope inspection window size $k$, Loss slope decay threshold $0 < \theta < 1$, Loss slope estimation error $e$, AdaLRS start and end step $t_{\text{start}},t_{\text{end}}$
}
\KwOut{Optimal learning rate $\eta^*$}
\SetKwFunction{FMain}{AdaLRS}
\SetKwProg{Fn}{Function}{}{}

\Fn{\FMain{$\eta_0$, $\alpha$, $\beta$, $k$, $\theta$, $t_{start}$, $t_{end}$}}{
    \textbf{Initialize:} $t \leftarrow 0$, $\eta_t \leftarrow \eta_0$, $\theta \leftarrow \theta_0$\;
    Initialize loss history $\mathcal{H} \leftarrow \emptyset$\;
    
    \While{$t_{\text{start}} < t < t_{\text{end}}$}{
        \tcp{\color{gray}1.Collect loss observations}
        Run $k$ training steps with $\{\eta_t,...,\eta_{t+k}\}$\; 
        Record losses $\{l_{t},...,l_{t+k}\}$\;
        Update $\mathcal{H} \leftarrow \mathcal{H} \cup \{l_t,...,l_{t+k}\}$\;
        
        \tcp{\color{gray}2.Calculate loss descent velocity}
        $v_t \leftarrow \mathrm{least\_square\_fit}(\{(1, l_t),...,(k, l_{t+k})\})$\;
        
        \tcp{\color{gray}3.Check velocity degradation}
        \If{$|\mathcal{H}| \geq 2k$ \textbf{and} $v_t < v_{t-k} \cdot \theta$}{
            \tcp{\color{gray}Trigger upscale phase}
            Run $k$ up-scaling steps with $\eta_{\text{temp}}$ gradually up-scaled to $\alpha\eta_t$\;
            
            \tcp{\color{gray}Find comparable historical records}
            Run another $k$ validation steps with loss records $\mathcal{L}_{\text{new}}=\{l_{t+2k+1},...,l_{t+3k}\}$\;
            $\mathcal{L}_{\text{ref}} \leftarrow \mathop{\mathrm{argmin}}\limits_{\mathcal{L} \subseteq \mathcal{H}} |\bar{\mathcal{L}} - \bar{\mathcal{L}_{\text{new}}}|$\;
            
            \tcp{Compare velocities}
            \If{$v_{\text{new}} > v_{\text{ref}} + 2e$}{
                $\eta_{t+k} \leftarrow \max(\alpha\lambda^t, 1)\eta_{t}$ \tcp*{\color{gray}Keep up-scaled LR}
            }
            \ElseIf{$v_{\text{new}} < v_{\text{ref}} - 2e$}{
                $\eta_{t+k} \leftarrow \frac{\eta_t}{\max(\beta\lambda^t, 1)}$ \tcp*{\color{gray}Downscale after failed upscale}
                Restore model states to step $t+k$\;
            }
            \Else{
                Restore model states to step $t+k$\;
            }
            $\mathcal{H},\theta \leftarrow \emptyset,\theta_0$\;
        }
        \ElseIf{$|\mathcal{H}| \geq 2k$ \textbf{and} $v_t,v_{t-k}<0$}{
            $\eta_{t+k} \leftarrow \frac{\eta_t}{\max(\beta\lambda^t, 1)}$ \tcp*{\color{gray}Downscale if loss rises consecutively}
            $\mathcal{H},\theta \leftarrow \emptyset,\theta_0$
            Restore model states to step $t+k$\;
        }
        \Else{
            $\eta_{t+k} \leftarrow \eta_t$ \tcp*{\color{gray}Maintain current LR}
            $\theta \leftarrow \frac{\theta+1}{2}$ \tcp*{\color{gray}Narrow the decay threshold}
        }
        $t \leftarrow t + k$\;
    }
    \Return $\eta_t$\;
}
\end{algorithm}

In the algorithm pseudo code, we omit the following details for presentation simplicity: 
1) the learning rate $\eta_t$ is updated by a base scheduler according to the training step $t$, multiplied by a scaling factor from AdaLRS; 
2) during learning rate up-scaling, an early stopping mechanism is triggered if the training loss elevates to be higher than the maximum loss value in loss history; 
3) for corner cases where $\mathop{\mathrm{min}}\limits_{\mathcal{L} \subseteq \mathcal{H}} |\bar{\mathcal{L}} - \bar{\mathcal{L}_{\text{new}}}|$ exceeds a certain threshold, LR up-scaling or down-scaling are triggered if the resulted training loss is higher or lower than all history records.

\section{Proof for the Validity of LR Scaling Factor Values}
\label{appendix:val}
In this section, we demonstrate the validity of the multiplicatively independent design of LR scaling factors. 

\begin{theorem}
\label{theorem:3}
Let $\alpha,\beta>1$ be multiplicatively independent real numbers (for all integers $m,n$, $\alpha^m=\beta^n \implies m=n=0$), for any positive rational number $r$ and $\epsilon>0$, there exist integers $m,n\geq0$, such that:
\begin{equation}
    |\alpha^m\beta^{-n}-r|<\epsilon. 
\end{equation}
\end{theorem}

We start by pointing out that $\frac{\ln \alpha}{\ln \beta}$ is irrational. 
This is rather straightforward because if $\frac{\ln \alpha}{\ln \beta}$ is rational, then $\alpha^q=\beta^p$, where $p,q$ are positive integers, which contradicts with the multiplicatively independent design, and thus $\frac{\ln \alpha}{\ln \beta}$ must be irrational.

Then we apply Kronecker's Theorem.
Let $\theta=\frac{\ln\alpha}{\ln\beta}$. 
For any real number $c$ and $\epsilon'>0$, there exist integers $m,n \in \mathbb{N}$ such that $\left| m\theta - n - c \right| < \epsilon'$.
Choose $c=\frac{\ln r}{\ln \beta}$. 
Then there exist $m,n \in \mathbb{N}$ satisfying:
\begin{equation}
    \left| m\frac{\ln \alpha}{\ln \beta} - n - \frac{\ln r}{\ln \beta} \right| < \epsilon'. 
\end{equation}
Multiplying through by $\ln\beta$, we get: 
\begin{equation}
\label{eq:get}
    \left| m\ln \alpha - n\ln \beta - \ln r \right| < \epsilon' \ln \beta.
\end{equation}

Finally, let $\epsilon'=\frac{\epsilon}{|q|\ln\beta}$, for sufficiently small $\epsilon'$, by the continuity of the exponential function:
\begin{equation}
    \left| \frac{\alpha^m}{\beta^n} - r \right| = |r| \cdot \left| e^{m\ln \alpha - n\ln \beta - \ln r} - 1 \right|.
\end{equation}
Using Equation~\ref{eq:get}, we bound:
\begin{equation}
    \left| \frac{\alpha^m}{\beta^n} - r \right| < |r| \cdot \epsilon' \ln \beta = \epsilon, 
\end{equation}
which proves Theorem~\ref{theorem:3}.

\section{Limitations}

\label{sec:limit}
Despite the promising optimal learning rate search effectiveness and performance improvement, AdaLRS fails to achieve comparable results with appropriate LR baselines with excessively large LR settings. 
As discussed in Section~\ref{sec:main_res_2} and Section~\ref{sec:exp_ct}, we attribute this problem to the disruptive parameter updates performed by large learning rates. 
This result restricts the application of AdaLRS on arbitrary LR settings. 
We recommend to apply AdaLRS with relatively small initial learning rates for optimal performance, which is validated to work smoothly in both from-scratch and continual pretraining, and leave this problem to be solved in future works. 

Moreover, the design of AdaLRS prioritizes the generalizability to unexplored foundation model pretraining tasks, and therefore we do not study the extent to which AdaLRS can approximate the optimal LR in this work. 
We will explore the impact of the scaling factor designs and learning rate search ranges to demonstrate the fine-grained convergence of AdaLRS in our future work. 

We also point out that the near-optimal learning rates used in Fit LR experiments in Section~\ref{exp} are obtained from grid search pilot experiments.
We search for the optimal LR setting with exponentially larger LR in a given range, such as $[2e-6,4e-6,8e-6,2e-5,...,8e-3]$ for LLM experiments. 
Although sufficiently precise for our experiments, the resulting approximated search result may introduce slight variance in optimal LR estimation.


\end{document}